\documentclass[  english%, ngerman				% Sprachen
               , bibliography = totoc			% Literaturverzeichnis ohne Nummer im Inhaltsverzeichnis
               , DIV = 12                    % Seiteneinteilung festsetzen, calc berechnet DIV=9 (scrguide S. 40)
               , parskip = false					% Definition Absatz (scrguide.pdf, Seite 83)
               , numbers = noenddot          % keine amerikanische Kapitelnummerierung Chapter 2.1. (enddot)
               , paper = a4                  % Papierformat
               , pagesize							% set pdf interna
               , toc = graduated             % Form und Inhalt des Inhaltsverzeichnisses (scrguide.pdf, Seite 78)
					, twoside = false             % einseitiges Dokument
              ]{scrartcl}						% Dokumentklasse: scrartcl, scrreprt, scrbook, scrlttr2
\usepackage[T1]{fontenc}
\usepackage[utf8]{inputenc}
\usepackage{  adjustbox								% adjust material in several ways
				, amsmath                        % ams-Mathematikpaket mit Summen- und Mengensymbolen
				, amssymb                        % erweiterter mathematischer Symbolsatz (Pfeile, etc.)
            , babel									% Sprachpacket
				, bbm                            % Mengensymbole im Gravurtyp, z.B. IR
				, blindtext
            , booktabs                       % Verwendung schöner Tabellenlinien (\toprule, \midrule, \bottomrule)
            , calc									% Berechnungen in TeX
				, chngcntr								% Counter nicht zurücksetzen (Fußnoten fortlaufend)
				, datatool								% externe Daten laden
				, enumitem								% Zählumgebung einfach ändern
				, eurosym								% EUR-Symbol
            , graphicx                       % erweitertes Einbinden von Graphiken
				, listliketab							% Tabellen sehen wie Listen aus
				, longtable								% Tabellen über mehrere Seiten (benoetigt fuer longtabu)
				, makecell								% 
				, mathtools								% enhance the appearance of documents containing a lot of math
				, multirow								% mehrere Zeilen zu einer zusammenfassen
            , nicefrac                       % schönere Brüche in mathematischen Umgebungen
            , pgfplots								% benutzerfreundlich Graphiken erstellen
				, setspace								% Durchschuss besser verwalten
				, tabu									% flexible Tabellen auch über mehrere Seiten (enthaelt tabular, tabular*, tabularx, array)
				, tikz									% Makropaket für pgf
				, translator							% Übersetzer
            , trfsigns								% definiert \im (imaginäre Einheit), \e (eulersche Zahl) und weitere
            , wasysym                       	% mehr Symbole
            , xcolor									% vielfältige Farbenspiele
			  }
\usepackage{mathpazo}
\usepackage{tgheros}
\usepackage[OT1, euler-digits]{eulervm}				% mathematics fonts based on Euler and CM (load before fontspec)
\usepackage{ifluatex}
\ifluatex
	\usepackage{fontspec}								% Spezifikation Zeichensatz
\fi
\usepackage[section]{placeins}
\usepackage[texindy]{imakeidx}
\usepackage[  backend = biber
				, style = authoryear
				, backref = true
				, url = true
				, dashed = false
				, maxbibnames = 99
			  ]{biblatex}
\usepackage[  autostyle = true
				, german = quotes	% sollte automatisch so sein
				, english = british
			  ]{csquotes}
\usepackage[  binary-units = true
				, detect-weight = true
				, detect-family = true
				, detect-shape = true
				, math-micro = \mu					% necessary for the use with eulervm
				, math-degree = {}^\circ
				, per-mode = fraction
				, separate-uncertainty = true		% Ungenauigkeit: 8.2(13) vs. 8.2+-1.3
				, table-number-alignment = center
				, retain-unity-mantissa = false
			  ]{siunitx}
\usepackage[  os = win
			  ]{menukeys}
\usepackage[normalem]{ulem}
\usepackage[  calcwidth = 0.9\linewidth		% Breite bei Benutzung ausrechnen
				, font = small							% alles kleiner
				, labelfont = bf						% Bezeichner und Trenner zusätzlich fett
				, hypcap = true						% Sprung zur Graphik nicht zum Label
			  ]{caption}
\usepackage{varioref}
\colorlet{mygreen}{green!50!black}
\colorlet{myblue}{blue!100!black}
\usepackage{hyperref}
\usepackage[  nameinlink
				, noabbrev
			  ]{cleveref}
\ifluatex
	\defaultfontfeatures{Ligatures=TeX}
\fi
\addto\extrasenglish{\sisetup{locale = UK}}
\usepackage[  automark
            , headsepline         			% Linie unter der Kopfzeile
            , footsepline						% Linie über der Fußzeile für scrheadings (normal Seite)
            , plainfootsepline				% Linie über der Fußzeile für scrplain (Kapitelanfänge, Verzeichnisse)
           ]{scrlayer-scrpage}
\clearscrheadfoot                   % löscht alle Elemente aus scrheadings und scrplain

	\automark[subsection]{section}	% setzt \rightmark[] auf subsection und \leftmark{} auf section
\ohead[]{\rightmark}               	% außen: plain: nix; normal: secion(kurz)überschrift
\ofoot[\translate{Page}\ \pagemark]{\translate{Page}\ \pagemark}	% außen: Seitenzahlen
\deffootnote{1em}{1em}{\textsuperscript{\thefootnotemark\ }}
\addtokomafont{pageheadfoot}{\normalfont}
\DeclareNameAlias{default}{last-first}
\AtBeginDocument{%
}

\makeatletter
	\let\getDocDate\@date
\makeatother

\pgfplotsset{compat=1.14}
\usetikzlibrary{babel, chains, positioning, scopes}
\pgfkeys{  /pgf/number format/.cd
			, set thousands separator={\,}
			, min exponent for 1000 sep=4
}
\usepgfplotslibrary{units, statistics}
\AtEveryBibitem{\clearfield{pagetotal}} % clears number of pages
\DeclareMathOperator{\E}{E}

\newcommand*{\R}{\texttt{R}}

\DeclareSIUnit\Umdr{\minute^{-1}}
\DeclareSIUnit\EURO{\text{\euro}}

\newcommand*{\showgrid}[3][5]{%
	\providecommand{\griddepth}{#1}%
	\resizebox{#2}{!}{%
		\begin{tikzpicture}[inner sep=0, rounded corners=0pt]
			\node[anchor=south west] (image) at (0, 0) {#3};
			\foreach \iThick in {0, ..., \griddepth} {%
				\path (image.north east) ++(-\iThick, -\iThick) coordinate(topright);
				\draw[semithick, red] (\iThick, \iThick) rectangle (topright);
				\ifnum\iThick<\griddepth
					\foreach \iThin in {1, ..., 4} {%
						\path (image.north east) ++(-\iThick, -\iThick) ++(-\iThin/5, -\iThin/5) coordinate(topright);
						\draw[very thin, green] (\iThick, \iThick) ++(\iThin/5, \iThin/5) rectangle (topright);
					}
				\fi
			}
		\end{tikzpicture}
	}
}

\setlist{nosep}

\hypersetup{  pdftitle = {Infill Criterion for Multimodal Model-Based Optimisation}
				, pdfauthor = {Dirk Surmann}
				, colorlinks = true
				, urlcolor = mygreen					% Farbe für Hyperlinks
				, urlbordercolor = mygreen			% Farbe des Rahmens für Hyperlinks
				, citecolor = mygreen				% Links zum Literaturverzeichnis
				, citebordercolor = mygreen		% Rahmen für Links zum Literaturverzeichnis
				, linkcolor = myblue					% Interne Links
				, linkbordercolor = myblue			% Rahmenfarbe interne Links
				, linktocpage = false				% Seitenzahlen(false) statt Text sind verlinkt
				, bookmarksopen = true				% Bookmarks werden geöffnet
				, bookmarksnumbered = true			% Bookmarks sind nummeriert
				, unicode = true						% use only unicode characters in pdf
}

\DeclareMathOperator{\EI}{EI}
\DeclareMathOperator{\GEILM}{GEILM}
\DeclareMathOperator{\PR}{PR}
\DeclareMathOperator{\AHD}{AHD}
\newcommand*{\xvec}{\boldsymbol{x}}
\newcommand*{\mnsm}{\hat{\mu}(\xvec)} % mean of surrogate model
\newcommand*{\sdsm}{\hat{s}(\xvec)} % standard deviation of surrogate model

\begin{document}
% select language: english or ngerman
\selectlanguage{english}
% define title page
\titlehead{{\Large TU Dortmund University\\}
			  Faculty of Statistics}
%\subject{Example}
\title{Infill Criterion for\\Multimodal Model-Based Optimisation}
%\subtitle{MISSING}
\author{Dirk Surmann\\Uwe Ligges\\Claus Weihs}
\date{\today}	% \today
%\publishers{}					% Herausgeber, Dozenten etc.

% numbering i, ii, ...
\pagenumbering{roman}

% display title page
\maketitle

% show list of fixme's (only in draft mode)
%\listoffixmes

% Verzeichnisse einbinden
%\tableofcontents							% Damit wird das Inhaltsverzeichnis eingebunden

%\noindent
%\begin{minipage}{\textwidth}
%	\listoffigures                      % Abbildungsverzeichnis
%	\listoftables                       % Tabellenverzeichnis
%\end{minipage}

% numbering 1, 2, ...
\pagenumbering{arabic}

% abstract
\begin{abstract}
Physical systems are modelled and investigated within simulation software in an increasing range of applications.
In reality an investigation of the system is often performed by empirical test scenarios which are related to typical situations.
Our aim is to derive a method which generates diverse test scenarios each representing a challenging situation for the corresponding physical system.

From a mathematical point of view challenging test scenarios correspond to local optima.
Hence, we focus to identify all local optima within mathematical functions.
Due to the fact that simulation runs are usually expensive we use the model-based optimisation approach with its well-known representative efficient global optimisation.
We derive an infill criterion which focuses on the identification of local optima.
The criterion is checked via fifteen different artificial functions in a computer experiment.
Our new infill criterion performs better in identifying local optima compared to the expected improvement infill criterion and Latin Hypercube Samples.
\end{abstract}

\section{Introduction}
\label{secIntroduction}

Simulation software is widely used in a variety of applications.
We mention two representative applications.
Power transmission systems are simulated to shed light on problematic situations and how to deal with them in practice \parencites{surmann_modelling_2014}{surmann_predicting_2017}.
Machine engineers check the behaviour of a new aircraft wing within simulation software before building it in reality.
In most of such complex simulations the application is tested by applying empirical test scenarios.
One possible scenario in a power network is a line fault in combination with an increased power consumption.
Does the transmission system handle this situation?
The aircraft wing is bent by a specific angle and checked for cracks after the test.
After passing different test scenarios successfully the application is deemed to be safe regarding all possible influences.
However, all of these test scenarios have different points in common.
Firstly, they are designed in a manual fashion.
Secondly, every test scenario should reflect a challenging situation.
Thirdly, the amount of test scenarios for an application should cover almost all possible situations.
Finally, a simulation run is expensive in the majority of cases.

Our goal is a quality improvement of test scenarios to generate diverse test data each representing a challenged situation for an application.
Due to the fact, that a simulation is expensive in its execution, we use the efficient global optimisation (EGO) approach proposed by \textcite{jones_efficient_1998}.
EGO is based on a measured response from a black box function with the objective to find its global optimum.
An exemplary response in a transmission system could be the simulated time until the simulation of a power network fails.
In the aircraft wing example we can use the number of cracks or their averaged length.
Multi-objective model-based optimisation \parencite{bischl_moimbo_2014} covers multiple responses.

EGO and model-based optimisation (MBO) mainly focus on global optimisation.
\Textcite{bischl_mlrmbo_2017} give an overview of state of the art techniques regarding multi-objective MBO using parallel computing.
An additional criterion for multi-objective optimisation is given by \textcite{bischl_moimbo_2014}.
The topic at hand obtains a set of good solutions in contrast to a single global optimum.
\Textcite{wessing_true_2017} discussed the search for multiple optima instead of a global one.
By using all samples from EGO rating them via topographical selection \parencite{torn_topographical_1992} the approach is capable to identify different optima.
In a computer experiment with twelve artificial test problems \textcite{wessing_true_2017} work out the differences between four algorithms.

The paper at hand aims to improve the efficient search of multiple optima in expensive functions.
For the sake of simplicity the focus is on local minima which can be switched to local maxima by inverting the corresponding function.
\Cref{secMethods} describes the used methods, especially \cref{subsecMethodsGEILM} provides a new infill criterion which aims to identify local minima.
We focus to find all minima of the corresponding function.
In the identification process minima with lower function values are more interesting than those with higher values.
The criterion is checked in \cref{secExperiment} via fifteen different test functions in a computer experiment.
To rate the results, we work out all local minima of the artificial test functions and list them in \cref{secAppMinima}.
\Cref{secConclusion} summarises the paper in a conclusion.

\section{Methods}
\label{secMethods}

This section describes the general MBO algorithm and one of its representatives, EGO, in \cref{subsecMethodsMBO}.
We introduce an infill criterion to identify local minima in \cref{subsecMethodsGEILM}.
The identification of local minima within the corresponding surrogate function is described in \cref{subsecMethodsIdentifyMinima}.
Rating the solutions of the different algorithms is specified in \cref{subsecMethodsRateSolutions}.

\subsection{Model-Based Optimisation}
\label{subsecMethodsMBO}

Let $f:\mathcal{X}\rightarrow\mathbb{R}$ be an arbitrary deterministic objective function with a $p$-dimensional numeric input domain $\mathcal{X}=\left[\boldsymbol{l},\boldsymbol{u}\right]\subset\mathbb{R}^p$.
The vectors $\boldsymbol{l}=\left(l_1, \dotsc, l_d\right)^\top$ and $\boldsymbol{u}=\left(u_1, \dotsc, u_d\right)^\top$ are the lower and upper bounds of $\mathcal{X}$, respectively.

The neighbourhood of a point $\xvec^\star\in\mathcal{X}$ is defined by $N(\xvec^\star)=\left\lbrace \xvec\in\mathcal{X}|d\left(\xvec, \xvec^\star\right)\leq\epsilon\right\rbrace$ with $\epsilon>0$ and a metric $d:\mathcal{X}\times\mathcal{X}\rightarrow\mathbb{R}^+$.
$f\left(\xvec^\star\right)$ is a local minimum if $\exists\epsilon>0\colon\nexists\xvec\in N\left(\xvec^\star\right)\colon f\left(x\right)<f\left(\xvec^\star\right)$.
As described by \textcite{wessing_true_2017} this definition ensures the global minimum to be a local minimum, even if it includes all plateaus.
All local minima are summarised in the solution set $S=\left\lbrace\xvec\in\mathcal{X}|f\left(\xvec\right)=f\left(x_i^\star\right)\right\rbrace$ where $f\left(\xvec_i^\star\right)$ are the $i=1,\dotsc,h$ local minima of the objective of the test function.
For the sake of simplicity, we are interested in all local minima of the given test function.
A definition to restrict the number of minima to a given value can be found in \textcite[sec. 2]{wessing_true_2017} as well as some brief ideas on additional constraints for the solution set.

Model-based optimisation (MBO) is usually used in an environment where $f$ is expensive to evaluate, hence only a limited number of function evaluations is allowed.
In every iteration $f$ is approximated via a much cheaper to evaluate surrogate model (or meta-model) $\hat{f}$.
The general MBO approach is outlined in the following list and is described in depth by \textcite{bischl_mlrmbo_2017}.
\begin{enumerate}
	\item Generate an initial design $D\subset\mathcal{X}$ (usually Latin Hypercube Desgin \parencite{mckay_comparison_1979}) and calculate $\boldsymbol{y}=f(D)$.
	\item The sequential phase starts fitting a surrogate model to the evaluated points $D$ and the corresponding values $\boldsymbol{y}$.
	\item Get additional point $\xvec'$ proposed by infill criterion (see \cref{subsecMethodsGEILM}).
			The criterion works on $\hat{f}$ and determine points which are promising for optimisation.
	\item Evaluate $\xvec'$ to $y'$ using $f$ and extend $D$ and $\boldsymbol{y}$, respectively.
	\item If no termination criteria are met (number of evaluations, etc.) go to step 2.
	\item Return $\hat{y}^\star=\min(\boldsymbol{y})$ and corresponding $\hat{\xvec}^\star$ as proposed global optimum for $f$.
\end{enumerate}
Step 2 of the MBO approach fits a surrogate model as a cheaper to evaluate function to the current design $D$ with respect to the evaluations $\boldsymbol{y}$.
The model choice has a main effect on the approximation of the objective function.
Because $\mathcal{X}\subset\mathbb{R}^d$ \textsc{Kriging} \parencites{jones_efficient_1998}{williams_gaussian_2006} is recommended and provides a direct estimation of the prediction standard error, or local uncertainty measure, next to the estimation of the true function value $f(\xvec)$.
For instance, EGO is a well known \textsc{Kriging} based approach.

\subsection{Gradient Enhanced Inspection of Local Minima}
\label{subsecMethodsGEILM}

The infill criterion is another essential part of MBO.
It leads the optimisation to handle the trade-off between exploitation and exploration by using a combination of different statistics from the surrogate model $\hat{f}$.
In most situations the estimators $\mnsm$ and $\sdsm$, estimated by $\hat{f}$ are used in a single formula to handle the trade-off in a well-balanced manner.
\Textcite{jones_efficient_1998} proposed the expected improvement $\EI(\xvec)$ as infill criterion, which is the most popular criterion and widely used.
It is defined as $\EI(\xvec)\coloneqq\E\left(\max\left\lbrace\hat{y}^\star-Y(\xvec), 0\right\rbrace\right)$, where $Y(\xvec)$ is a random variable that expresses the posterior distribution at $\xvec$, estimated via the surrogate model $\hat{f}$.
Using a \textsc{Kriging} model $Y(\xvec)$ is normally distributed with $Y(\xvec)\sim\mathcal{N}\left(\mnsm,\hat{s}^2(\xvec)\right)$.
Using this assumption, $\EI(\xvec)$ can be expressed as
\begin{align}
	\label{eqnMethodsEI}
	\EI(\xvec) &= \left(\hat{y}^\star-\mnsm\right)
						\Phi\left(\frac{\hat{y}^\star-\mnsm}{\sdsm}\right)+
						\sdsm\phi\left(\frac{\hat{y}^\star-\mnsm}{\sdsm}\right)
\end{align}
where $\phi$ and $\Phi$ are the density and distribution function of the standard normal distribution, respectively.
In the first addend the difference between the current minimum $\hat{y}^\star$ and the local estimator $\mnsm$ is rated high for lower values of $\mnsm$.
Its corresponding standard deviation $\sdsm$ is rated high for its higher values.
Hence, the expected improvement leads us to points with low $\mnsm$ and high $\sdsm$.

Simpler approaches to compound $\mnsm$ and $\sdsm$ for a point $\xvec$ are given by the lower confidence bound ($\operatorname{LCB}(\xvec)$) or the standard error ($\operatorname{SE}(\xvec)$) itself in \cref{eqnMethodsSimpleInfill}, where $\lambda>0$ is a constant to control the trade-off between both estimators in \cref{eqnMethodsLCB}.
\begin{subequations}
	\label{eqnMethodsSimpleInfill}
	\begin{align}
		\label{eqnMethodsLCB}
		\operatorname{LCB}(\xvec;\lambda) &= \mnsm - \lambda\sdsm\\
		\label{eqnMethodsSE}
		\operatorname{SE}(\xvec) &= \sdsm
	\end{align}
\end{subequations}
The infill criterion $\operatorname{SE}(\xvec)$ in \cref{eqnMethodsSE} handles the trade-off by shifting the whole weight to exploration.
It will try to cover the design space of the objective function equally, to reduce the local standard error.

Our purpose to inspect local minima yields a modified form of $\EI(\xvec)$ and $\operatorname{SE}(\xvec)$.
We search for meaningful test scenarios to challenge the corresponding application.
This usually results in lower values of the response function compared to non-challenging test scenarios.
A meaningful test scenario is translated into the model-based optimisation as a local minimum whose function value is near to the global minimum.
The slope at a point between two local minima has to be significantly different from \num{0}, otherwise it is a plateau which is considered as one minimum.
Hence, we enhance the new infill criterion to skip exploration in regions with a steep surrogate function.
The gradient enhanced inspection of local minima $\GEILM(\xvec)$, as new infill criterion, is defined by
\begin{align}
	\label{eqnMethodsGEILM}
	\GEILM(\xvec) &= \sdsm\Phi\left(\frac{\hat{y}^\star-\mnsm}{s_\mathrm{p}}\right)
							g_\lambda\left(\left\lVert\nabla\mnsm\right\rVert_{\infty}\right)
\end{align}
with
\begin{align}
	\label{eqnMethodsSEquant}
	s_\mathrm{p} &= \min\left\lbrace s\in\mathbb{R}^+\middle|
			\frac{\hat{y}^\star-\max(\boldsymbol{y})}{s}=\Phi^{-1}(p)
		\right\rbrace
\end{align}
where $\Phi$ is the distribution function of the standard normal distribution.
$g_\lambda$ is the density of the exponential distribution with parameter $\lambda$.
$\lVert\cdot\rVert_{\infty}$ describes the supremum norm of $\cdot$ and is called maximum norm in case of a vector $\boldsymbol{a}=\left(a_1,\dotsc,a_n\right)^\top$.
In this case it takes the form $\lVert\boldsymbol{a}\rVert_{\infty}=\max\left\lbrace\lvert a_1\rvert,\dotsc,\lvert a_n\rvert\right\rbrace$.
The differential operator, or nabla operator, $\nabla$ is defined in terms of partial derivative operators and denotes the gradient of a scalar field.
We choose $\operatorname{SE}(\xvec)$ with its explorative nature as a starting point and use the multiplication operator to implement a weighting on $\operatorname{SE}(\xvec)$.
Summing up the coefficients is only meaningful for infill criteria which deal with $\mnsm$ directly and do not cover $\mnsm$ in a function, as shown in \cref{eqnMethodsEI,eqnMethodsLCB}, because $\mnsm$ is related to exploitation whereas $\sdsm$ is related to exploration.
The weighting via $\Phi$ is reused from $\EI(\xvec)$ to add a connection to the expected function value $\mnsm$.
$\Phi$ weights down $\mnsm$ the more it differs from the current minimum $\hat{y}^\star$ which reflects the higher interest in local minima with lower function values.
We adjust the standardisation via a $p$-quantile standard deviation $s_p$ with $p\in(0,1)$.
$s_p$ is driven by the range of evaluated design points and independent of $\sdsm$.
This approach supports $\GEILM(\xvec)$ exploring the design space with higher expected values $\mnsm$ with a lower priority.
Due to the fact that a gradient at a point between two optima is significantly different from \num{0} we add a second weighting via the exponential distribution $g_\lambda$.
It considers the maximum partial derivative to ensure the highest weighting for local optima.
Design points outside local optima (or plateaus) get lower priority which further enhances $\GEILM(\xvec)$ to use the available number of runs in promising regions.

Two exemplary design points illustrate the behaviour of $\GEILM$.
In the global minimum $\hat{y}^\star$ the total weight on $\sdsm$ evaluates to its maximum of $\frac{1}{2}\lambda$.
In this case $\Phi$ evaluates to $\frac{1}{2}$ and $g_\lambda$ to $\lambda$ because the gradient in an optimum is \num{0}.
Hence, $\GEILM(\xvec)$ evaluates to $\frac{1}{2}\lambda\sdsm$ for the global minimum.
If we assume an example point which is not a local minimum, with a high value of the local estimator compared to the global minimum, $\GEILM(\xvec)$ converges to \num{0}.
The function $\Phi$ converges to \num{0} for higher values of $\mnsm$.
The function $g_\lambda$ converges to \num{0} for increasing absolute values of the gradient.
If the surrogate function is unexplored at this point $\sdsm$ uprates the infill criterion $\GEILM(\xvec)$.

\subsection{Identify Local Minima of Surrogate Function}
\label{subsecMethodsIdentifyMinima}

Identifying the local minima after performing the model-based optimisation is done via topographical selection \parencite{torn_topographical_1992} by \textcite{wessing_true_2017}.
We follow a different approach and instead use the surrogate function of evaluated points from the objective function to identify the local minima.
This method has the advantage to detect minima in areas with lower exploitation.
However, the surrogate function can estimate the objective function inaccurately.
To identify the local minima, we draw a Latin Hypercube Sample \parencite{stein_large_1987} $U^\star\subset\mathcal{X}$ of size $n=200^{\log_{3}\left(p+3-1\right)}$ according to the number of dimensions $p$.
The formula for $n$ is defined empirically to balance the expected workload and required points covering the input domain $\mathcal{X}$ in \cref{subsecMethodsMBO}.
We apply the quasi-\textsc{Newton} algorithm with box constraints (L-BFGS-B) defined by \textcite{byrd_limited_1995} and described by \textcite{nocedal_numerical_2006} to all points $\xvec\in U^\star$.
This gradient descent algorithm moves each point $\xvec$ to its next local minimum $\xvec_i^\star$ of the surrogate function $\hat{f}$ with $i=1,\dotsc,k$ local minima.
We skip all points in $U^\star$ which moved to the limits of $\mathcal{X}$ assuming the local minima of the objective function within the input domain.
The approximation set $U$ is defined by agglomerations near the representatives in $U^\star$ via
\begin{align}
	\label{eqnMethodsApproxSet}
	U=\left\lbrace\xvec\in\mathcal{X}\vert d_{\mathrm{Che}}\left(\xvec,U^\star\right)\leq\delta\right\rbrace
\end{align}
with $\delta>0$.
The function $d_{\mathrm{Che}}\left(\xvec,U^\star\right)$ denotes the Chebyshev distance \parencite{abello_handbook_2002} of a point $\xvec$ to its nearest neighbour in $U^\star$.

Different approaches to identify multiple optima were applied in the field of multimodal optimisation.
\Textcite{gudla_automated_2005} proposed a hybrid between genetic and gradient algorithms.
Evolutionary algorithms are used by \textcite{deb_finding_2010} in a concept of multi-objective optimisation to solve a single-objective problem.
\Textcite{stoean_multimodal_2010} combine their genetic algorithm with a topological separation of subpopulations and apply the approach to different test functions.
A hybrid of \textsc{Nelder}-\textsc{Mead} algorithm and gradient descent method is applied by \textcite{abbas_method_2018} to signal processing.
Our approach spreading points in the input domain via a Latin Hypercube Sample and move them to their nearest local minima is simple and empowers us to calculate large samples.
Further research will show if more complex algorithms identify local optima considerably faster.

\subsection{Rate Solution Set}
\label{subsecMethodsRateSolutions}

Global optimisation uses the deviation from the global optimum as a performance measure.
In multimodal optimisation we use the number of found local minima $l=\left\lvert U\right\rvert$ divided by the correct number of optima $\left\lvert S\right\rvert=h$ as peak ratio $\PR(U)=\frac{l}{h}$.
Because the surrogate function is used to identify the local minima we define the peak ratio different than \textcite{ursem_multinational_1999}.
They choose the set of local minima $S$ as a reference set to check if each minimum is met by a point in the final design $D$.
We deal with $\mathcal{X}$ to identify the agglomerated number of found optima in $U^\star$.
Hence, $\PR(U)\in\left(0,\infty\right)$ which reveals improper fits of the surrogate functions for high values of $\PR$.

Another measure is the averaged \textsc{Hausdorff} distance ($\AHD$) described by \textcites{hausdorff_mengenlehre_1927}{rockafellar_variational_2004}.
It is defined by
\begin{align}
	\label{eqnMethodsAHD}
	\AHD(U)=\max\left\lbrace
						\left(\frac{1}{l}\sum_{i=1}^{l}d_\mathrm{nn}(\xvec_i, U)^r\right)^\frac{1}{r},
						\left(\frac{1}{h}\sum_{i=1}^{h}d_\mathrm{nn}(\xvec_i^\star, S)^r\right)^\frac{1}{r}
					\right\rbrace
\end{align}
using $S$ as a reference set.
The function $d_\mathrm{nn}(\xvec,X)$ denotes the Euclidean distance of a point $\xvec$ to a its nearest neighbour in a set of points $X$.

\section{Experiment}
\label{secExperiment}

This section describes the computer experiment to analyse the proposed infill criterion with its setup in \cref{subsecExpSetup}.
\Cref{subsecExpExtreme} analysis the extreme values occur in the results.
A comparison between the three methods is discussed in \cref{subsecExpCriteria}.

\subsection{Setup}
\label{subsecExpSetup}

We evaluate the performance of the infill criterion $\GEILM$ using an extensive computer experiment.
It is compared to the most popular $\EI$ infill criterion and a Latin Hypercube Sample \parencite{stein_large_1987} using $n_\mathrm{of}=15$ artificial objective functions as black box functions.
\Cref{tabExpFunctions} contains the objective functions used in this experiment.
\begin{table}[htbp]
	\centering
	\caption{Objective Functions used for Testing}
	\label{tabExpFunctions}
	\sisetup{round-mode = places, round-precision = 3, table-format = 3.0e0}
	\begin{tabular}{lS[table-format = 1.0e0]Slr}
		\toprule
		\thead{Function Name} & {\thead{Dim.}} & \multicolumn{2}{c}{\thead{\#Local Minima}} & \thead{Reference}\\
		\midrule
		Alpine Function No. 02 & 1 & 2 & \Cref{tabAppMinimaAlpine02.1} & \textcite{clerc_swarm_1999}\\
		Alpine Function No. 02 & 2 & 5 & \Cref{tabAppMinimaAlpine02.2} & \textcite{clerc_swarm_1999}\\
		Alpine Function No. 02 & 3 & 14 & \Cref{tabAppMinimaAlpine02.3} & \textcite{clerc_swarm_1999}\\
		\textsc{Branin} Function & 2 & 3 & \Cref{tabAppMinimaBranin} & \textcite{dixon_global_1978}\\
		Cosine Mixture Function & 1 & 5 & \Cref{tabAppMinimaCosineMix.1} & \AtNextCite{\defcounter{maxnames}{1}}\textcite{ali_numerical_2005}\\
		Cosine Mixture Function & 2 & 25 & \Cref{tabAppMinimaCosineMix.2} & \AtNextCite{\defcounter{maxnames}{1}}\textcite{ali_numerical_2005}\\
		Cosine Mixture Function & 3 & 125 & \Cref{tabAppMinimaCosineMix.3} & \AtNextCite{\defcounter{maxnames}{1}}\textcite{ali_numerical_2005}\\
		\textsc{Hartmann} Function & 3 & 3 & \Cref{tabAppMinimaHartmann.3} & \textcite{dixon_global_1978}\\
		\textsc{Hartmann} Function & 6 & 2 & \Cref{tabAppMinimaHartmann.6} & \textcite{dixon_global_1978}\\
		\textsc{Himmelblau} Function & 2 & 4 & \Cref{tabAppMinimaHimmelblau} & \textcite{himmelblau_applied_1972}\\
		Modified \textsc{Rastrigin} Function & 4 & 48 & \Cref{tabAppMinimaModRastrigin.4} & \textcite{deb_multimodal_2012}\\
		Modified \textsc{Rastrigin} Function & 8 & 48 & \Cref{tabAppMinimaModRastrigin.8} & \textcite{deb_multimodal_2012}\\
		\textsc{Shekel} Function $5$ & 4 & 5 & \Cref{tabAppMinimaShekel.5} & \textcite{dixon_global_1978}\\
		\textsc{Shekel} Function $7$ & 4 & 7 & \Cref{tabAppMinimaShekel.7} & \textcite{dixon_global_1978}\\
		\textsc{Shekel} Function $10$ & 4 & 10 & \Cref{tabAppMinimaShekel.10} & \textcite{dixon_global_1978}\\
		\bottomrule
	\end{tabular}
\end{table}
We list the name, the dimensionality, the number of local minima, and a reference for each objective function.
All local minima are worked out in a computationally intensive task and tabulated in \cref{secAppMinima}, because only data of the global minimum can be found in the literature.
The set of objective functions consist of the classic test set for global optimisation by \textcite{dixon_global_1978}, which is a subset of test problems described by \textcite{ali_numerical_2005}.
We expand this set by the Alpine Function No. 02 \parencite{clerc_swarm_1999}, the Cosine Mixture Function \parencite{ali_numerical_2005}, the \textsc{Himmelblau} Function \parencite{himmelblau_applied_1972} and the modified \textsc{Rastrigin} Function \parencite{deb_multimodal_2012}.

Model-based optimisation is split into an initial and a sequential phase (\cref{subsecMethodsMBO}).
The number of points drawn in the initial phase is equal to $n_{\mathrm{init}}=\left\lbrace3^2, 4^2, \dotsc, 8^2\right\rbrace^\top$.
In the sequential stage $n_{\mathrm{seq}}=\left\lbrace3^2, 4^2, \dotsc, 12^2\right\rbrace^\top$ are added using the infill criterion.
All elements in both vectors are squared, to focus on results with a lower number of design points.
Designs with higher numbers are less interesting, because the real world simulations are too expensive to evaluate a high amount of design points.
All combinations of elements in $n_\mathrm{init}$ and $n_\mathrm{seq}$ are evaluated in the experiment.
We evaluate $\left\lvert n_\mathrm{init}\right\rvert\cdot\left\lvert n_\mathrm{seq}\right\rvert\cdot n_\mathrm{of}=900$ experiments for each infill criterion.
To compare the infill criteria to a Latin Hypercube Sample (LHS), we fit the same \textsc{Kriging} model as in MBO to a LHS with $n_\mathrm{LHS}=\left\lbrace4^2, 5^2, \dotsc, 15^2\right\rbrace^\top$ design points which results in $\left\lvert n_\mathrm{LHS}\right\rvert\cdot n_\mathrm{of}=180$ experiments.
Each experiment is repeated \num{30} times which results in $\left(2\cdot 900+180\right)\cdot 30=\num{59400}$ runs, including infill criteria $\EI$ and $\GEILM$.
We set the hyperparameters of $\GEILM$ criterion in \cref{eqnMethodsGEILM,eqnMethodsSEquant} to $\lambda=2$ and $p=0.001$, respectively.
Both choices were made empirically to uprate low gradients ($\lambda$) and rate down extreme values ($p$).
Further research can be done on these hyperparameters.

In each run we record two performance measurements to rate the solution set (\cref{subsecMethodsRateSolutions}).
The peak ratio ($\PR$) which considers representatives for every local minimum in \cref{eqnMethodsApproxSet} with $\delta=0.001$ and the averaged \textsc{Hausdorff} distance ($\AHD$) with $r=1$ in \cref{eqnMethodsAHD}.
All implementation is done in \R\ \parencite{rdevelopmentcoreteam_language_2017} via package \texttt{mlrMBO} \parencite{bischl_mlrmbo_2017}.
We parallelise the experiment using package \texttt{batchtools} \parencite{lang_batchtools_2017}.

\subsection{Extreme Values}
\label{subsecExpExtreme}

The construction of $\PR$ results in the interval $(0,\infty)$.
However, we expect two soft limits \num{0} and \num{1}.
No minima are found with $\PR=0$ whereas a value of \num{1} indicates that all local minima are found.
Due to construction, $\PR$ is capable to detect overfitted surrogate models.
If the \textsc{Kriging} model finds too many optima, $\PR$ exceeds \num{1} and indicates overfitting by higher values.
Hence, we are interested in a reasonable dataset which covers the expected interval $[0,1]$ as well as the overfitting indication.
We choose the interval $[0,5]$ empirically.
The value $\PR=5$ indicates a surrogate function with five times more minima than the corresponding objective function and indicates overfitting.

\begin{figure}[htbp]
	\centering
	\includegraphics[trim=0mm 0mm 0mm 0mm, clip, width=\linewidth]{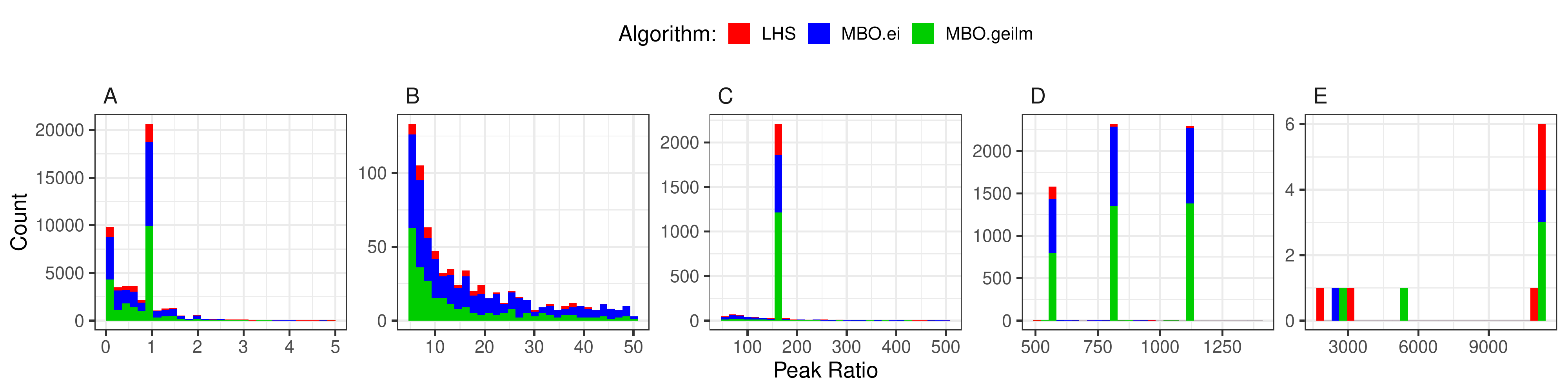}
	\caption{Extreme Values in $\PR$ Intervals by Algorithm}
	\label{figExpExtremePRbyIntervalAlgo}
\end{figure}

A first analysis of $\PR$ shows extreme values up to \num{11260.5} and points out \SI{16.3}{\percent} runs with $\PR > 5$ as shown in \cref{tabExpExtremePRbyInterval}.
\begin{table}[htbp]
	\centering
	\caption{Number of Extreme Values in $\PR$ Intervals}
	\label{tabExpExtremePRbyInterval}
	\sisetup{round-mode = places, table-format = 5.0e0}
	\begin{tabular}{lS}
		\toprule
		\thead{Intervall} & {\thead{Count}}\\
		\midrule
		$\mathrm{A}=[0,5]$ & 49697 \\
		$\mathrm{B}=(5,50]$ & 734 \\ 
		$\mathrm{C}=(50, 500]$ & 2698 \\ 
		$\mathrm{D}=(500, 1500]$ & 6259 \\ 
		$\mathrm{E}=(1500, \infty)$ & 12 \\
		\bottomrule
	\end{tabular}
\end{table}
The table lists the count of $\PR$ values in the corresponding intervals.
We choose these remaining interval ranges for B to E in a way to plot values of $\PR$ in histograms which illustrate the extreme values.
The different information shown in \cref{figExpExtremePRbyIntervalAlgo} will be hidden if the data is plotted in one histogram.
Extreme values are independent of the used method.
Certainly, the ratio of LHS algorithm is lower, because the number of chosen MBO combinations is ten times higher.
The histogram A shows two peaks indicating the soft limits \num{0} and \num{1} of $\PR$.
Higher values indicate a misleading surrogate function with too many local minima.
Histogram B illustrates the exponential decrease for higher values of $\PR$ which can be observed in histogram A, too.
The peaks in histograms C, D and E are far outside this decreasing histogram range described by A and B and calculated in \cref{tabExpExtremePRbyProb}.
\begin{table}[htbp]
	\centering
	\caption{Number of Extreme Values in $\PR$ by Interval and Problem}
	\label{tabExpExtremePRbyProb}
	\sisetup{round-mode = places, table-format = 5.0e0}
	\begin{tabular}{lSSSSS}
		\toprule
		& \multicolumn{5}{c}{\textbf{Interval Counts}}\\
		\thead{Function} & {\thead{B}} & {\thead{C}} & {\thead{D}} & {\thead{E}} & {\thead{B -- E}}\\
		\midrule
		Alpine02.2 & 3 & 130 & & & 133 \\ 
		Alpine02.3 & 79 & 2087 & & & 2166 \\ 
		CosineMix.3 & 4 & & & & 4 \\ 
		Hartmann.3 & & & 1 & & 1 \\ 
		Hartmann.6 & 26 & 12 & 6 & 12 & 56 \\ 
		Himmelblau & & 2 & & & 2 \\ 
		modRastrigin.4 & 2 & 2 & & & 4 \\ 
		modRastrigin.8 & 2 & 3 & 1 & & 6 \\ 
		Shekel.5 & 175 & 135 & 2332 & & 2642 \\ 
		Shekel.7 & 154 & 152 & 2339 & & 2645 \\ 
		Shekel.10 & 289 & 175 & 1580 & & 2044 \\ 
		\bottomrule
	\end{tabular}
\end{table}
Apparently, the Alpine Function No. 02 and the \textsc{Shekel} Functions 5, 7, and 10 seem to be difficult to fit, they generate nearly all extreme values.
A closer look does not reveal any relation to the number of design points, neither the number of initial points nor the number of sequential points.
The divergence in these functions is spread through the complete experimental space.
Because interval B contains only functions discovered in C to E, we decide to refer the further analysis to interval A.

\subsection{Comparison of Infill Criteria}
\label{subsecExpCriteria}

\begin{figure}[htbp]
	\centering
	\includegraphics[trim=0mm 0mm 0mm 0mm, clip, width=\linewidth]{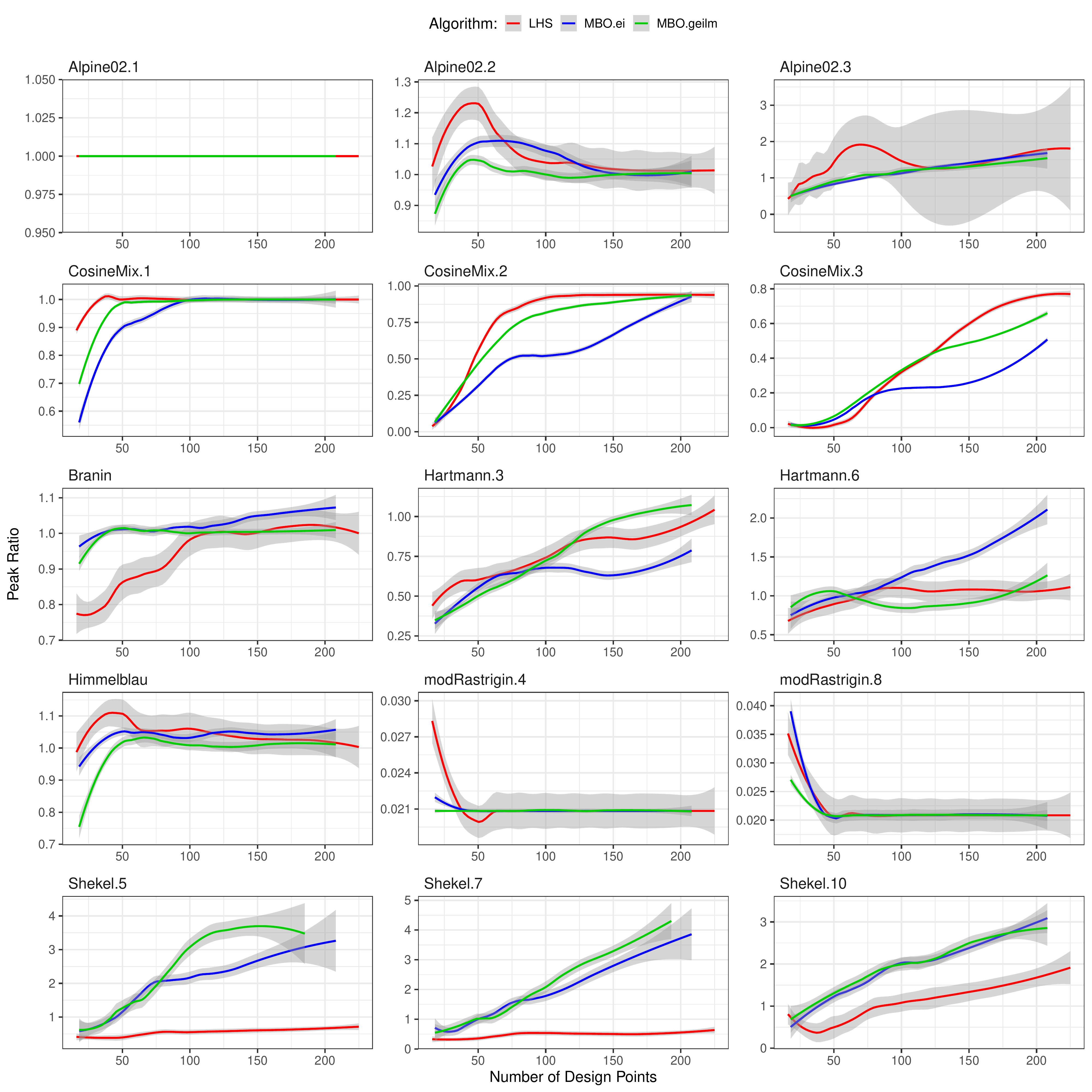}
	\caption{Peak Ratio ($\PR$) over Number of Design Points Grouped by Algorithm}
	\label{figExpPR}
\end{figure}

\begin{figure}[htbp]
	\centering
	\includegraphics[page=2, trim=0mm 0mm 0mm 0mm, clip, width=\linewidth]{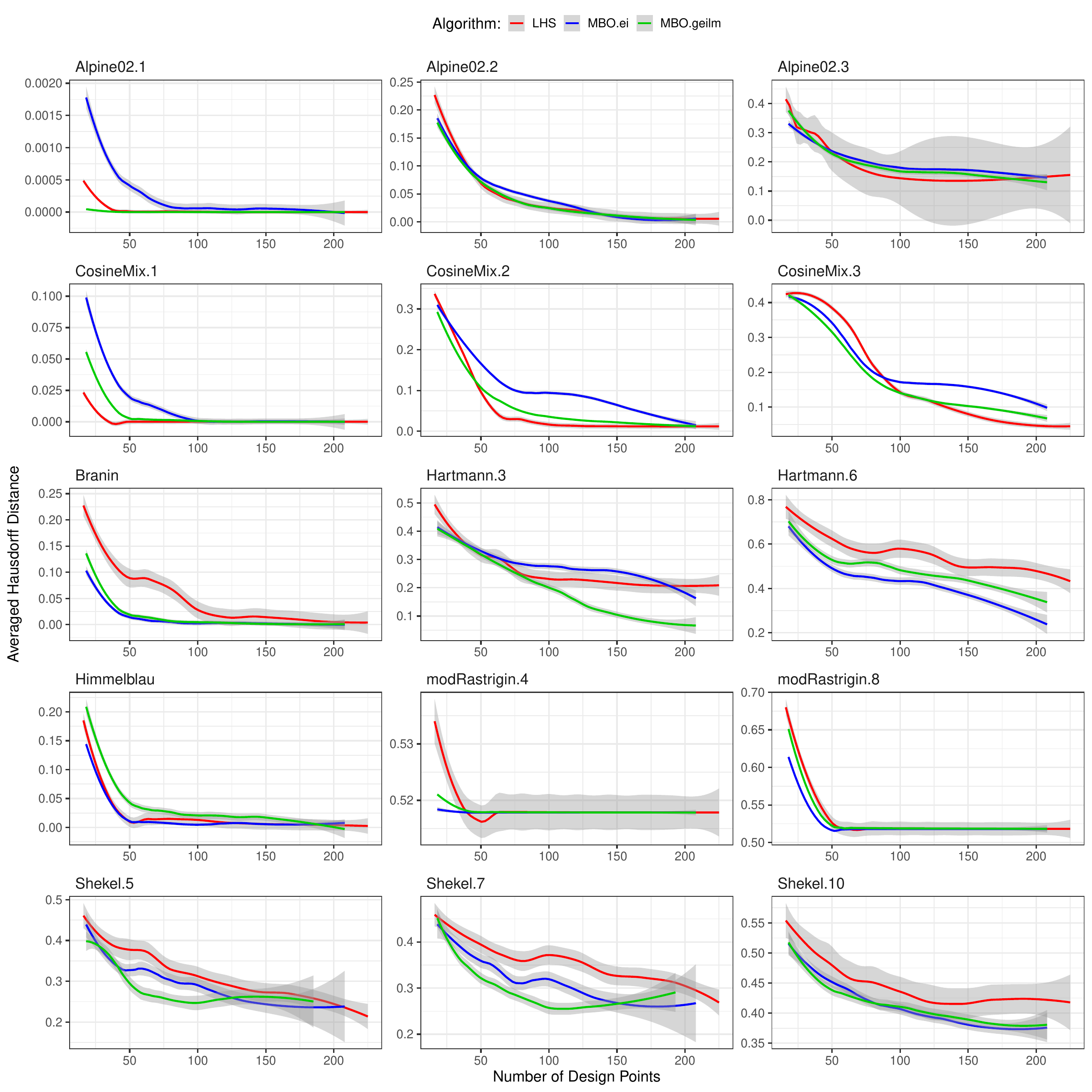}
	\caption{Averaged \textsc{Hausdorff} distances ($\AHD$) over number of design points grouped by algorithm}
	\label{figExpAHD}
\end{figure}

\Cref{figExpPR,figExpAHD} illustrate the progress of the two performance measurements over the number of design points.
In all figures, the curves represent a local regression (LOESS) for smoothing points in scatterplots described in \textcite{cleveland_local_2017} with a smoothing parameter $\alpha=\num{0.5}$.
The parameter $\alpha$ controls the fraction of the neighbouring data points which are used to fit each local polynomial.
Each curve is shown with a grey \SI{95}{\percent} confidence interval for the smoothed value assuming a normal distribution.

For most objective functions the proposed infill criterion $\GEILM$ beats $\EI$ in terms of peak ratio ($\PR$), shown in \cref{figExpPR}.
It converges earlier to the optimal value of \num{1} which can be seen especially for Alpine Function No. 02, Cosine Mixture Function and \textsc{Hartmann} Function.
Especially, \textsc{Hartmann} Function in \num{6} dimensions shows the need for an infill criterion designed to identify local minima.
$\EI$ diverges over the number of design points contrary to the proposed criterion which identifies in average the correct number of local minima.
This example illustrates that $\EI$ identifies the global optimum, as expected.
\textsc{Branin}, \textsc{Hartmann}, and \textsc{Himmelblau} Function illustrates the high variance of LHS.
The smoothed mean of LHS outperforms MBO in many objective functions.
However, the confidence intervals are often much wider compared to MBO methods.
Additionally, LHS has plateaus in the \textsc{Branin} Function or overshoots in the \textsc{Himmelblau} Function and Alpine Function No. 02 in dimensionalities \num{2} and \num{3}.
\Cref{figExpAHD} shows a similar picture for the averaged \textsc{Hausdorff} distance.
We reach the optimal value of $\AHD=0$ with $\GEILM$ continuously faster than by using $\EI$ or LHS.
Latin Hypercube Sampling is sometimes better in terms of mean $\AHD$ but shows a higher variance compared to the MBO approach.

All three methods are highly overstrained with the objective functions Modified \textsc{Rastrigin} and \textsc{Shekel}.
The latter one was already identified in \cref{subsecExpExtreme}, where it generates about $\frac{3}{4}$ of the extreme values.
Whereas $\AHD$ results in lesser values with more design points, we see the divergence of $\PR$ in \cref{figExpPR} for all \textsc{Shekel} Functions.
For the modified \textsc{Rastrigin} Function low values of $\PR$ illustrate the overstrained behaviour to identify the local minima.
Only \num{1} of \num{48} local minima is identified for this type of objective functions.
We suppose a problematic fit of \textsc{Kriging} models to a modified \textsc{Rastrigin} Function.
From our point of view, the dimension or number of local minima is not the only cause for these results, since the Cosine Mixture Function in \num{3} dimensions with \num{125} local minima shows a good performance in $\PR$ and $\AHD$ as well as \textsc{Hartmann} Function in \num{6} dimension with \num{2} local optima.

\begin{figure}[htbp]
	\centering
	\includegraphics[trim=0mm 0mm 0mm 0mm, clip, width=\linewidth]{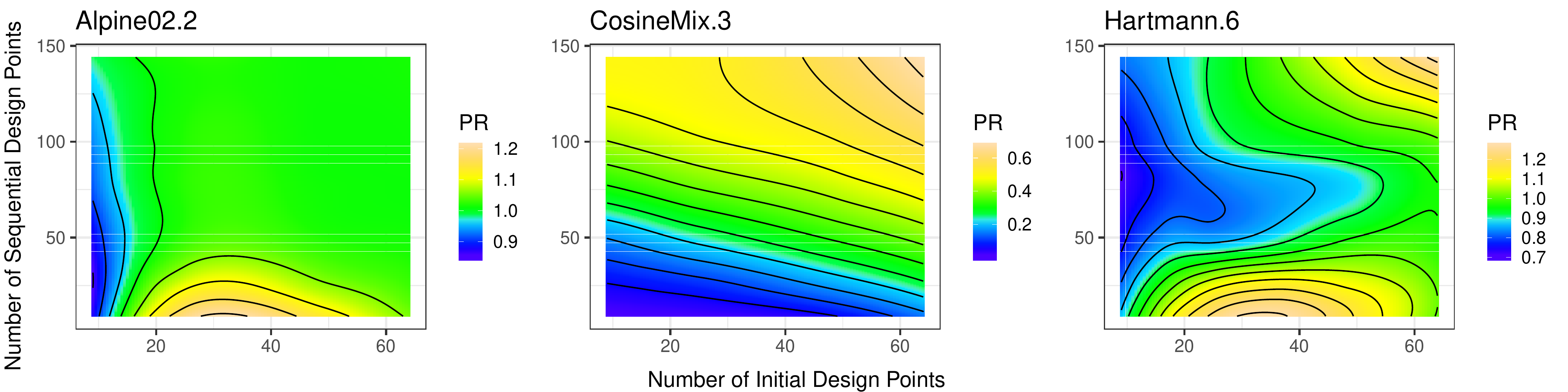}
	\caption{Contour Plot of Peak Ratio ($\PR$) over Number of Initial and Sequential Design Points for Infill Criterion $\GEILM$.}
	\label{figExpGEILMPR}
\end{figure}

In case of MBO, we are interested in the development of performance measurements over the number of initial and sequential design points.
Contour plots of peak ratio over the number of initial and sequential design points, smoothed by LOESS, are shown in \cref{figExpGEILMPR} for infill criterion $\GEILM$.
The contour plot of averaged \textsc{Hausdorff} distance is shown in appendix \cref{figAppGEILMAHD}.
We chose Alpine Function No. 02 in \num{2} dimensions, Cosine Mixture Function in \num{3} dimensions and \textsc{Hartmann} Function in \num{6} dimensions as showcases because the corresponding plots illustrate the variety of contours between the different objective functions.
Contour plots of both performance measurements for all objective functions are given in appendix \cref{figAppGEILMPRall,figAppGEILMAHDall}.
Local minima of the \num{2} dimensional function Alpine02 are found fast with a low number of initial and sequential points.
Interestingly, a higher number of points in the initial design results in $\PR>1$ which indicates an overfitted \textsc{Kriging} model.
The overfit is resolved in the sequential stage with a low number of design points.
An undersized initial design is hard to compensate because MBO needs a high number of design points in the sequential stage to achieve $\PR=1$.
Cosine Mixture Function in \num{3} dimensions is an example of an expected contour plot for $\PR$.
A higher number of points in the initial or sequential design results in a constant improvement of peak ratio.
A modified behaviour of Alpine02 can be found in the contour plot of \textsc{Hartmann} Function in \num{6} dimensions.
We see an overfitted surrogate model with \num{35} initial design points and the minimum number of points in the sequential stage.
The same reaction holds for the highest amount of initial and design points which was observed in \cref{figExpPR} as an increase in the curve of $\GEILM$ for a high number of design points.
Medium number of points in the sequential stage sometimes result in an incorrect number of local minima from the \textsc{Kriging} model.
This has a strong impact in LOESS because the peak ratio for a function with \num{2} local minima has a precision of \num{0.5}.
The corresponding contour plot of averaged \textsc{Hausdorff} distance in \cref{figAppGEILMAHD} underlines this statement.
Improvements of $\AHD$ for increasing number of points in the initial and sequential stage are illustrated by contour plots for each objective function (see \cref{figAppGEILMAHDall}).

\section{Conclusion}
\label{secConclusion}

We introduce the infill criterion gradient enhanced inspection of local minima ($\GEILM$) for model-based optimisation (MBO) which aims to identify local minima in expensive objective functions.
It is capable to identify all local minima and focuses on minima with lower values of the objective function.
Minima with lower values are explored more intense than minima with higher function values.
A computer experiment compares the behaviour of $\GEILM$ to the most popular infill criterion expected improvement ($\EI$) used in efficient global optimisation (EGO).
Additionally, we include Latin Hypercube Sampling (LHS) which reflects a state of the art design for computer experiments not considering the objective function.
We work out and tabulate all local minima of the used objective functions.
A variety of objective functions is tested and shows a good performance of $\GEILM$ in case of the measurements peak ratio ($\PR$) and averaged \textsc{Hausdorff} distance ($\AHD$).
It outperforms $\EI$ and LHS in the averaged performance measurements and especially the latter in terms of variance.
The need for special infill criteria to identify local minima instead of the global minimum is obvious by considering the \textsc{Hartmann} Function in \num{6} dimensions.
$\PR$ diverges using EGO over an increasing number of design points contrary to $\GEILM$ which identifies the correct number of local optima.
From our point of view, it is worth the effort to define criteria purpose-built for dealing with local minima.

Future work should examine the hyperparameters of $\GEILM$ and improve the criterion to minimise the number of hyperparameters.
Additionally, we may investigate the used surrogate model in MBO because of its poor behaviour with respect to the objective functions Alpine Function No. 02 and \textsc{Shekel} Functions 5, 7, and 10.
Finally, the infill criteria and used surrogate model should be tested in a larger computer experiment with additional objective functions especially of higher dimensions ($p\geq 6$) and a greater variety of function types.
This will clarify the discussed issues and make stronger statements about the capabilities of the infill criteria aiming at local minima.

\section{Acknowledgement}

This work has been funded by the Deutsche Forschungsgemeinschaft (DFG), Forschergruppe 1511 \textit{Schutz- und Leitsysteme zur zuverlässigen und sicheren elektrischen Energieübertragung}.

\clearpage
\appendix

\section{Local Minima}
\label{secAppMinima}

This section lists all local minima of the used objective functions.
The rows are ordered according to the function value $y$ beginning with the global minima.

\subsection{Alpine Function No. 02}

\begin{table}[htbp]
	\centering
	\caption{Local Minima of \num{1}-Dimensional Alpine02}
	\label{tabAppMinimaAlpine02.1}
	\sisetup{round-mode = places, round-precision = 3, table-format = -1.3e0}
	\begin{tabular}{rSS}
		\toprule
		\thead{\#} & \multicolumn{1}{c}{$\xvec$} & {\thead{$y$}}\\
		\midrule
		1 & 7.917 & -2.808 \\
		2 & 1.837 & -1.308 \\
		\bottomrule
	\end{tabular}
\end{table}

\begin{table}[htbp]
	\centering
	\caption{Local Minima of \num{2}-Dimensional Alpine02}
	\label{tabAppMinimaAlpine02.2}
	\sisetup{round-mode = places, round-precision = 3, table-format = -1.3e0}
	\begin{tabular}{rSSS}
		\toprule
		\thead{\#} & \multicolumn{2}{c}{$\xvec$} & {\thead{$y$}}\\
		\midrule
		1 & 7.917 & 7.917 & -7.886 \\ 
		2 & 4.816 & 4.816 & -4.764 \\ 
		3 & 1.837 & 7.917 & -3.672 \\ 
		4 & 7.917 & 1.837 & -3.672 \\ 
		5 & 1.837 & 1.837 & -1.710 \\ 
		\bottomrule
	\end{tabular}
\end{table}

\begin{table}[htbp]
	\centering
	\caption{Local Minima of \num{3}-Dimensional Alpine02}
	\label{tabAppMinimaAlpine02.3}
	\sisetup{round-mode = places, round-precision = 3, table-format = -1.3e0}
	\begin{tabular}{rSSSS[table-format = -2.3e0]}
		\toprule
		\thead{\#} & \multicolumn{3}{c}{$\xvec$} & {\thead{$y$}}\\
		\midrule
		1  & 7.917 & 7.917 & 7.917 & -22.144 \\ 
		2  & 4.816 & 4.816 & 7.917 & -13.379 \\ 
		3  & 4.816 & 7.917 & 4.816 & -13.379 \\ 
		4  & 7.917 & 4.816 & 4.816 & -13.379 \\ 
		5  & 1.837 & 7.917 & 7.917 & -10.311 \\ 
		6  & 7.917 & 1.837 & 7.917 & -10.311 \\ 
		7  & 7.917 & 7.917 & 1.837 & -10.311 \\ 
		8  & 1.837 & 4.816 & 4.816 & -6.230 \\ 
		9  & 4.816 & 1.837 & 4.816 & -6.230 \\ 
		10 & 4.816 & 4.816 & 1.837 & -6.230 \\ 
		11 & 1.837 & 1.837 & 7.917 & -4.802 \\ 
		12 & 1.837 & 7.917 & 1.837 & -4.802 \\ 
		13 & 7.917 & 1.837 & 1.837 & -4.802 \\ 
		14 & 1.837 & 1.837 & 1.837 & -2.236 \\
		\bottomrule
	\end{tabular}
\end{table}

\FloatBarrier
\subsection{\textsc{Branin} Function}

\begin{table}[htbp]
	\centering
	\caption{Local Minima of \num{2}-Dimensional \textsc{Branin}}
	\label{tabAppMinimaBranin}
	\sisetup{round-mode = places, round-precision = 3, table-format = -1.3e0}
	\begin{tabular}{rSSS}
		\toprule
		\thead{\#} & \multicolumn{2}{c}{$\xvec$} & {\thead{$y$}}\\
		\midrule
		1 & -3.142 & 12.275 & 0.398 \\ 
		2 & 3.142 & 2.275 & 0.398 \\ 
		3 & 9.425 & 2.475 & 0.398 \\ 
		\bottomrule
	\end{tabular}
\end{table}

\FloatBarrier
\subsection{Cosine Mixture Function}

\begin{table}[htbp]
	\centering
	\caption{Local Minima of \num{1}-Dimensional CosineMix}
	\label{tabAppMinimaCosineMix.1}
	\sisetup{round-mode = places, round-precision = 3, table-format = -1.3e0}
	\begin{tabular}{rSS}
		\toprule
		\thead{\#} & \multicolumn{1}{c}{$\xvec$} & {\thead{$y$}}\\
		\midrule
		1 & 0.000 & -0.100 \\ 
		2 & -0.369 & 0.048 \\ 
		3 & 0.369 & 0.048 \\ 
		4 & -0.725 & 0.487 \\ 
		5 & 0.725 & 0.487 \\
		\bottomrule
	\end{tabular}
\end{table}

\begin{table}[htbp]
	\centering
	\caption{Local Minima of \num{2}-Dimensional CosineMix}
	\label{tabAppMinimaCosineMix.2}
	\sisetup{round-mode = places, round-precision = 3, table-format = -1.3e0}
	\begin{tabular}{rSSS}
		\toprule
		\thead{\#} & \multicolumn{2}{c}{$\xvec$} & {\thead{$y$}}\\
		\midrule
		1  & 0.000 & 0.000 & -0.200 \\ 
		2  & -0.369 & 0.000 & -0.052 \\ 
		3  & 0.000 & -0.369 & -0.052 \\ 
		4  & 0.000 & 0.369 & -0.052 \\ 
		5  & 0.369 & 0.000 & -0.052 \\ 
		6  & -0.369 & -0.369 & 0.096 \\ 
		7  & -0.369 & 0.369 & 0.096 \\ 
		8  & 0.369 & -0.369 & 0.096 \\ 
		9  & 0.369 & 0.369 & 0.096 \\ 
		10 & -0.725 & 0.000 & 0.387 \\ 
		11 & 0.000 & -0.725 & 0.387 \\ 
		12 & 0.000 & 0.725 & 0.387 \\ 
		13 & 0.725 & 0.000 & 0.387 \\ 
		14 & -0.725 & -0.369 & 0.535 \\ 
		15 & -0.725 & 0.369 & 0.535 \\ 
		16 & -0.369 & -0.725 & 0.535 \\ 
		17 & -0.369 & 0.725 & 0.535 \\ 
		18 & 0.369 & -0.725 & 0.535 \\ 
		19 & 0.369 & 0.725 & 0.535 \\ 
		20 & 0.725 & -0.369 & 0.535 \\ 
		21 & 0.725 & 0.369 & 0.535 \\ 
		22 & -0.725 & -0.725 & 0.975 \\ 
		23 & -0.725 & 0.725 & 0.975 \\ 
		24 & 0.725 & -0.725 & 0.975 \\ 
		25 & 0.725 & 0.725 & 0.975 \\ 
		\bottomrule
	\end{tabular}
\end{table}

\begin{@empty}
	\sisetup{round-mode = places, round-precision = 3, table-format = -1.3e0}
	\begin{longtable}{rSSSS}
		\caption{Local Minima of \num{3}-Dimensional CosineMix}
		\label{tabAppMinimaCosineMix.3}\\
		\toprule
		\thead{\#} & \multicolumn{3}{c}{$\xvec$} & {\thead{$y$}}\\
		\midrule
		\endfirsthead
		\midrule
		\multicolumn{5}{c}{--- continued from previous page ---}\\
		\thead{\#} & \multicolumn{3}{c}{$\xvec$} & {\thead{$y$}}\\
		\midrule
		\endhead
		\multicolumn{5}{c}{--- continue on next page ---}\\
		\midrule
		\endfoot
		\bottomrule
		\endlastfoot
		1   & 0.000 & 0.000 & 0.000 & -0.300 \\ 
		2   & -0.369 & 0.000 & 0.000 & -0.152 \\ 
		3   & 0.000 & -0.369 & 0.000 & -0.152 \\ 
		4   & 0.000 & 0.000 & -0.369 & -0.152 \\ 
		5   & 0.000 & 0.000 & 0.369 & -0.152 \\ 
		6   & 0.000 & 0.369 & 0.000 & -0.152 \\ 
		7   & 0.369 & 0.000 & 0.000 & -0.152 \\ 
		8   & -0.369 & -0.369 & 0.000 & -0.004 \\ 
		9   & -0.369 & 0.000 & -0.369 & -0.004 \\ 
		10  & -0.369 & 0.000 & 0.369 & -0.004 \\ 
		11  & -0.369 & 0.369 & 0.000 & -0.004 \\ 
		12  & 0.000 & -0.369 & -0.369 & -0.004 \\ 
		13  & 0.000 & -0.369 & 0.369 & -0.004 \\ 
		14  & 0.000 & 0.369 & -0.369 & -0.004 \\ 
		15  & 0.000 & 0.369 & 0.369 & -0.004 \\ 
		16  & 0.369 & -0.369 & 0.000 & -0.004 \\ 
		17  & 0.369 & 0.000 & -0.369 & -0.004 \\ 
		18  & 0.369 & 0.000 & 0.369 & -0.004 \\ 
		19  & 0.369 & 0.369 & 0.000 & -0.004 \\ 
		20  & -0.369 & -0.369 & -0.369 & 0.143 \\ 
		21  & -0.369 & -0.369 & 0.369 & 0.143 \\ 
		22  & -0.369 & 0.369 & -0.369 & 0.143 \\ 
		23  & -0.369 & 0.369 & 0.369 & 0.143 \\ 
		24  & 0.369 & -0.369 & -0.369 & 0.143 \\ 
		25  & 0.369 & -0.369 & 0.369 & 0.143 \\ 
		26  & 0.369 & 0.369 & -0.369 & 0.143 \\ 
		27  & 0.369 & 0.369 & 0.369 & 0.143 \\ 
		28  & -0.725 & 0.000 & 0.000 & 0.287 \\ 
		29  & 0.000 & -0.725 & 0.000 & 0.287 \\ 
		30  & 0.000 & 0.000 & -0.725 & 0.287 \\ 
		31  & 0.000 & 0.000 & 0.725 & 0.287 \\ 
		32  & 0.000 & 0.725 & 0.000 & 0.287 \\ 
		33  & 0.725 & 0.000 & 0.000 & 0.287 \\ 
		34  & -0.725 & -0.369 & 0.000 & 0.435 \\ 
		35  & -0.725 & 0.000 & -0.369 & 0.435 \\ 
		36  & -0.725 & 0.000 & 0.369 & 0.435 \\ 
		37  & -0.725 & 0.369 & 0.000 & 0.435 \\ 
		38  & -0.369 & -0.725 & 0.000 & 0.435 \\ 
		39  & -0.369 & 0.000 & -0.725 & 0.435 \\ 
		40  & -0.369 & 0.000 & 0.725 & 0.435 \\ 
		41  & -0.369 & 0.725 & 0.000 & 0.435 \\ 
		42  & 0.000 & -0.725 & -0.369 & 0.435 \\ 
		43  & 0.000 & -0.725 & 0.369 & 0.435 \\ 
		44  & 0.000 & -0.369 & -0.725 & 0.435 \\ 
		45  & 0.000 & -0.369 & 0.725 & 0.435 \\ 
		46  & 0.000 & 0.369 & -0.725 & 0.435 \\ 
		47  & 0.000 & 0.369 & 0.725 & 0.435 \\ 
		48  & 0.000 & 0.725 & -0.369 & 0.435 \\ 
		49  & 0.000 & 0.725 & 0.369 & 0.435 \\ 
		50  & 0.369 & -0.725 & 0.000 & 0.435 \\ 
		51  & 0.369 & 0.000 & -0.725 & 0.435 \\ 
		52  & 0.369 & 0.000 & 0.725 & 0.435 \\ 
		53  & 0.369 & 0.725 & 0.000 & 0.435 \\ 
		54  & 0.725 & -0.369 & 0.000 & 0.435 \\ 
		55  & 0.725 & 0.000 & -0.369 & 0.435 \\ 
		56  & 0.725 & 0.000 & 0.369 & 0.435 \\ 
		57  & 0.725 & 0.369 & 0.000 & 0.435 \\ 
		58  & -0.725 & -0.369 & -0.369 & 0.583 \\ 
		59  & -0.725 & -0.369 & 0.369 & 0.583 \\ 
		60  & -0.725 & 0.369 & -0.369 & 0.583 \\ 
		61  & -0.725 & 0.369 & 0.369 & 0.583 \\ 
		62  & -0.369 & -0.725 & -0.369 & 0.583 \\ 
		63  & -0.369 & -0.725 & 0.369 & 0.583 \\ 
		64  & -0.369 & -0.369 & -0.725 & 0.583 \\ 
		65  & -0.369 & -0.369 & 0.725 & 0.583 \\ 
		66  & -0.369 & 0.369 & -0.725 & 0.583 \\ 
		67  & -0.369 & 0.369 & 0.725 & 0.583 \\ 
		68  & -0.369 & 0.725 & -0.369 & 0.583 \\ 
		69  & -0.369 & 0.725 & 0.369 & 0.583 \\ 
		70  & 0.369 & -0.725 & -0.369 & 0.583 \\ 
		71  & 0.369 & -0.725 & 0.369 & 0.583 \\ 
		72  & 0.369 & -0.369 & -0.725 & 0.583 \\ 
		73  & 0.369 & -0.369 & 0.725 & 0.583 \\ 
		74  & 0.369 & 0.369 & -0.725 & 0.583 \\ 
		75  & 0.369 & 0.369 & 0.725 & 0.583 \\ 
		76  & 0.369 & 0.725 & -0.369 & 0.583 \\ 
		77  & 0.369 & 0.725 & 0.369 & 0.583 \\ 
		78  & 0.725 & -0.369 & -0.369 & 0.583 \\ 
		79  & 0.725 & -0.369 & 0.369 & 0.583 \\ 
		80  & 0.725 & 0.369 & -0.369 & 0.583 \\ 
		81  & 0.725 & 0.369 & 0.369 & 0.583 \\ 
		82  & -0.725 & -0.725 & 0.000 & 0.875 \\ 
		83  & -0.725 & 0.000 & -0.725 & 0.875 \\ 
		84  & -0.725 & 0.000 & 0.725 & 0.875 \\ 
		85  & -0.725 & 0.725 & 0.000 & 0.875 \\ 
		86  & 0.000 & -0.725 & -0.725 & 0.875 \\ 
		87  & 0.000 & -0.725 & 0.725 & 0.875 \\ 
		88  & 0.000 & 0.725 & -0.725 & 0.875 \\ 
		89  & 0.000 & 0.725 & 0.725 & 0.875 \\ 
		90  & 0.725 & -0.725 & 0.000 & 0.875 \\ 
		91  & 0.725 & 0.000 & -0.725 & 0.875 \\ 
		92  & 0.725 & 0.000 & 0.725 & 0.875 \\ 
		93  & 0.725 & 0.725 & 0.000 & 0.875 \\ 
		94  & -0.725 & -0.725 & -0.369 & 1.022 \\ 
		95  & -0.725 & -0.725 & 0.369 & 1.022 \\ 
		96  & -0.725 & -0.369 & -0.725 & 1.022 \\ 
		97  & -0.725 & -0.369 & 0.725 & 1.022 \\ 
		98  & -0.725 & 0.369 & -0.725 & 1.022 \\ 
		99  & -0.725 & 0.369 & 0.725 & 1.022 \\ 
		100 & -0.725 & 0.725 & -0.369 & 1.022 \\ 
		101 & -0.725 & 0.725 & 0.369 & 1.022 \\ 
		102 & -0.369 & -0.725 & -0.725 & 1.022 \\ 
		103 & -0.369 & -0.725 & 0.725 & 1.022 \\ 
		104 & -0.369 & 0.725 & -0.725 & 1.022 \\ 
		105 & -0.369 & 0.725 & 0.725 & 1.022 \\ 
		106 & 0.369 & -0.725 & -0.725 & 1.022 \\ 
		107 & 0.369 & -0.725 & 0.725 & 1.022 \\ 
		108 & 0.369 & 0.725 & -0.725 & 1.022 \\ 
		109 & 0.369 & 0.725 & 0.725 & 1.022 \\ 
		110 & 0.725 & -0.725 & -0.369 & 1.022 \\ 
		111 & 0.725 & -0.725 & 0.369 & 1.022 \\ 
		112 & 0.725 & -0.369 & -0.725 & 1.022 \\ 
		113 & 0.725 & -0.369 & 0.725 & 1.022 \\ 
		114 & 0.725 & 0.369 & -0.725 & 1.022 \\ 
		115 & 0.725 & 0.369 & 0.725 & 1.022 \\ 
		116 & 0.725 & 0.725 & -0.369 & 1.022 \\ 
		117 & 0.725 & 0.725 & 0.369 & 1.022 \\ 
		118 & -0.725 & -0.725 & -0.725 & 1.462 \\ 
		119 & -0.725 & -0.725 & 0.725 & 1.462 \\ 
		120 & -0.725 & 0.725 & -0.725 & 1.462 \\ 
		121 & -0.725 & 0.725 & 0.725 & 1.462 \\ 
		122 & 0.725 & -0.725 & -0.725 & 1.462 \\ 
		123 & 0.725 & -0.725 & 0.725 & 1.462 \\ 
		124 & 0.725 & 0.725 & -0.725 & 1.462 \\ 
		125 & 0.725 & 0.725 & 0.725 & 1.462 \\ 
	\end{longtable}
\end{@empty}

\FloatBarrier
\subsection{\textsc{Hartmann} Function}

\begin{table}[htbp]
	\centering
	\caption{Local Minima of \num{3}-Dimensional \textsc{Hartmann}}
	\label{tabAppMinimaHartmann.3}
	\sisetup{round-mode = places, round-precision = 3, table-format = -1.3e0}
	\begin{tabular}{rSSSS}
		\toprule
		\thead{\#} & \multicolumn{3}{c}{$\xvec$} & {\thead{$y$}}\\
		\midrule
		1 & 0.115 & 0.556 & 0.852 & -3.863 \\ 
		2 & 0.109 & 0.860 & 0.564 & -3.090 \\ 
		3 & 0.369 & 0.118 & 0.268 & -1.001 \\ 
		\bottomrule
	\end{tabular}
\end{table}

\begin{table}[htbp]
	\centering
	\caption{Local Minima of \num{6}-Dimensional \textsc{Hartmann}}
	\label{tabAppMinimaHartmann.6}
	\sisetup{round-mode = places, round-precision = 3, table-format = -1.3e0}
	\begin{tabular}{rSSSSSSS}
		\toprule
		\thead{\#} & \multicolumn{6}{c}{$\xvec$} & {\thead{$y$}}\\
		\midrule
		1 & 0.202 & 0.150 & 0.477 & 0.275 & 0.312 & 0.657 & -3.322 \\ 
		2 & 0.405 & 0.882 & 0.846 & 0.574 & 0.139 & 0.038 & -3.203 \\ 
		\bottomrule
	\end{tabular}
\end{table}

\FloatBarrier
\subsection{\textsc{Himmelblau} Function}

\begin{table}[htbp]
	\centering
	\caption{Local Minima of \num{2}-Dimensional \textsc{Himmelblau}}
	\label{tabAppMinimaHimmelblau}
	\sisetup{round-mode = places, round-precision = 3, table-format = -1.3e0}
	\begin{tabular}{rSSS}
		\toprule
		\thead{\#} & \multicolumn{2}{c}{$\xvec$} & {\thead{$y$}}\\
		\midrule
		1 & -3.779 & -3.283 & 0.000 \\ 
		2 & -2.805 & 3.131 & 0.000 \\ 
		3 & 3.000 & 2.000 & 0.000 \\ 
		4 & 3.584 & -1.848 & 0.000 \\ 
		\bottomrule
	\end{tabular}
\end{table}

\FloatBarrier
\subsection{Modified \textsc{Rastrigin} Function}

\begin{@empty}
	\sisetup{round-mode = places, round-precision = 3, table-format = -1.3e0}
	\begin{longtable}{rSSSSS[table-format = -2.3e0]}
		\caption{Local Minima of \num{4}-Dimensional modified \textsc{Rastrigin}}
		\label{tabAppMinimaModRastrigin.4}\\
		\toprule
		\thead{\#} & \multicolumn{4}{c}{$\xvec$} & {\thead{$y$}}\\
		\midrule
		\endfirsthead
		\midrule
		\multicolumn{6}{c}{--- continued from previous page ---}\\
		\thead{\#} & \multicolumn{4}{c}{$\xvec$} & {\thead{$y$}}\\
		\midrule
		\endhead
		\multicolumn{6}{c}{--- continue on next page ---}\\
		\midrule
		\endfoot
		\bottomrule
		\endlastfoot
		1  & 0.249 & 0.249 & 0.166 & 0.125 & 0.789 \\ 
		2  & 0.249 & 0.249 & 0.166 & 0.374 & 1.786 \\ 
		3  & 0.249 & 0.249 & 0.498 & 0.125 & 2.118 \\ 
		4  & 0.249 & 0.746 & 0.166 & 0.125 & 2.779 \\ 
		5  & 0.746 & 0.249 & 0.166 & 0.125 & 2.779 \\ 
		6  & 0.249 & 0.249 & 0.498 & 0.374 & 3.115 \\ 
		7  & 0.249 & 0.746 & 0.166 & 0.374 & 3.776 \\ 
		8  & 0.746 & 0.249 & 0.166 & 0.374 & 3.776 \\ 
		9  & 0.249 & 0.249 & 0.166 & 0.623 & 3.781 \\ 
		10 & 0.249 & 0.746 & 0.498 & 0.125 & 4.108 \\ 
		11 & 0.746 & 0.249 & 0.498 & 0.125 & 4.108 \\ 
		12 & 0.746 & 0.746 & 0.166 & 0.125 & 4.769 \\ 
		13 & 0.249 & 0.249 & 0.830 & 0.125 & 4.776 \\ 
		14 & 0.249 & 0.746 & 0.498 & 0.374 & 5.105 \\ 
		15 & 0.746 & 0.249 & 0.498 & 0.374 & 5.105 \\ 
		16 & 0.249 & 0.249 & 0.498 & 0.623 & 5.110 \\ 
		17 & 0.746 & 0.746 & 0.166 & 0.374 & 5.766 \\ 
		18 & 0.249 & 0.746 & 0.166 & 0.623 & 5.771 \\ 
		19 & 0.746 & 0.249 & 0.166 & 0.623 & 5.771 \\ 
		20 & 0.249 & 0.249 & 0.830 & 0.374 & 5.773 \\ 
		21 & 0.746 & 0.746 & 0.498 & 0.125 & 6.098 \\ 
		22 & 0.249 & 0.746 & 0.830 & 0.125 & 6.766 \\ 
		23 & 0.746 & 0.249 & 0.830 & 0.125 & 6.766 \\ 
		24 & 0.249 & 0.249 & 0.166 & 0.873 & 6.773 \\ 
		25 & 0.746 & 0.746 & 0.498 & 0.374 & 7.095 \\ 
		26 & 0.249 & 0.746 & 0.498 & 0.623 & 7.100 \\ 
		27 & 0.746 & 0.249 & 0.498 & 0.623 & 7.100 \\ 
		28 & 0.746 & 0.746 & 0.166 & 0.623 & 7.761 \\ 
		29 & 0.249 & 0.746 & 0.830 & 0.374 & 7.763 \\ 
		30 & 0.746 & 0.249 & 0.830 & 0.374 & 7.763 \\ 
		31 & 0.249 & 0.249 & 0.830 & 0.623 & 7.768 \\ 
		32 & 0.249 & 0.249 & 0.498 & 0.873 & 8.102 \\ 
		33 & 0.746 & 0.746 & 0.830 & 0.125 & 8.756 \\ 
		34 & 0.249 & 0.746 & 0.166 & 0.873 & 8.763 \\ 
		35 & 0.746 & 0.249 & 0.166 & 0.873 & 8.763 \\ 
		36 & 0.746 & 0.746 & 0.498 & 0.623 & 9.090 \\ 
		37 & 0.746 & 0.746 & 0.830 & 0.374 & 9.753 \\ 
		38 & 0.249 & 0.746 & 0.830 & 0.623 & 9.758 \\ 
		39 & 0.746 & 0.249 & 0.830 & 0.623 & 9.758 \\ 
		40 & 0.249 & 0.746 & 0.498 & 0.873 & 10.092 \\ 
		41 & 0.746 & 0.249 & 0.498 & 0.873 & 10.092 \\ 
		42 & 0.746 & 0.746 & 0.166 & 0.873 & 10.753 \\ 
		43 & 0.249 & 0.249 & 0.830 & 0.873 & 10.760 \\ 
		44 & 0.746 & 0.746 & 0.830 & 0.623 & 11.748 \\ 
		45 & 0.746 & 0.746 & 0.498 & 0.873 & 12.082 \\ 
		46 & 0.249 & 0.746 & 0.830 & 0.873 & 12.750 \\ 
		47 & 0.746 & 0.249 & 0.830 & 0.873 & 12.750 \\ 
		48 & 0.746 & 0.746 & 0.830 & 0.873 & 14.740 \\ 
	\end{longtable}
\end{@empty}

\begin{@empty}
	\sisetup{round-mode = places, round-precision = 3, table-format = -1.3e0}
	\begin{longtable}{rSSSSSSSSS[table-format = -2.3e0]}
		\caption{Local Minima of \num{8}-Dimensional modified \textsc{Rastrigin}}
		\label{tabAppMinimaModRastrigin.8}\\
		\toprule
		\thead{\#} & \multicolumn{8}{c}{$\xvec$} & {\thead{$y$}}\\
		\midrule
		\endfirsthead
		\midrule
		\multicolumn{10}{c}{--- continued from previous page ---}\\
		\thead{\#} & \multicolumn{8}{c}{$\xvec$} & {\thead{$y$}}\\
		\midrule
		\endhead
		\multicolumn{10}{c}{--- continue on next page ---}\\
		\midrule
		\endfoot
		\bottomrule
		\endlastfoot
		1  & 0.495 & 0.249 & 0.495 & 0.249 & 0.495 & 0.166 & 0.495 & 0.125 & 2.769 \\ 
		2  & 0.495 & 0.249 & 0.495 & 0.249 & 0.495 & 0.166 & 0.495 & 0.374 & 3.766 \\ 
		3  & 0.495 & 0.249 & 0.495 & 0.249 & 0.495 & 0.498 & 0.495 & 0.125 & 4.098 \\ 
		4  & 0.495 & 0.249 & 0.495 & 0.746 & 0.495 & 0.166 & 0.495 & 0.125 & 4.759 \\ 
		5  & 0.495 & 0.746 & 0.495 & 0.249 & 0.495 & 0.166 & 0.495 & 0.125 & 4.759 \\ 
		6  & 0.495 & 0.249 & 0.495 & 0.249 & 0.495 & 0.498 & 0.495 & 0.374 & 5.095 \\ 
		7  & 0.495 & 0.249 & 0.495 & 0.746 & 0.495 & 0.166 & 0.495 & 0.374 & 5.756 \\ 
		8  & 0.495 & 0.746 & 0.495 & 0.249 & 0.495 & 0.166 & 0.495 & 0.374 & 5.756 \\ 
		9  & 0.495 & 0.249 & 0.495 & 0.249 & 0.495 & 0.166 & 0.495 & 0.623 & 5.761 \\ 
		10 & 0.495 & 0.249 & 0.495 & 0.746 & 0.495 & 0.498 & 0.495 & 0.125 & 6.088 \\ 
		11 & 0.495 & 0.746 & 0.495 & 0.249 & 0.495 & 0.498 & 0.495 & 0.125 & 6.088 \\ 
		12 & 0.495 & 0.746 & 0.495 & 0.746 & 0.495 & 0.166 & 0.495 & 0.125 & 6.748 \\ 
		13 & 0.495 & 0.249 & 0.495 & 0.249 & 0.495 & 0.830 & 0.495 & 0.125 & 6.756 \\ 
		14 & 0.495 & 0.249 & 0.495 & 0.746 & 0.495 & 0.498 & 0.495 & 0.374 & 7.085 \\ 
		15 & 0.495 & 0.746 & 0.495 & 0.249 & 0.495 & 0.498 & 0.495 & 0.374 & 7.085 \\ 
		16 & 0.495 & 0.249 & 0.495 & 0.249 & 0.495 & 0.498 & 0.495 & 0.623 & 7.090 \\ 
		17 & 0.495 & 0.746 & 0.495 & 0.746 & 0.495 & 0.166 & 0.495 & 0.374 & 7.746 \\ 
		18 & 0.495 & 0.249 & 0.495 & 0.746 & 0.495 & 0.166 & 0.495 & 0.623 & 7.751 \\ 
		19 & 0.495 & 0.746 & 0.495 & 0.249 & 0.495 & 0.166 & 0.495 & 0.623 & 7.751 \\ 
		20 & 0.495 & 0.249 & 0.495 & 0.249 & 0.495 & 0.830 & 0.495 & 0.374 & 7.753 \\ 
		21 & 0.495 & 0.746 & 0.495 & 0.746 & 0.495 & 0.498 & 0.495 & 0.125 & 8.077 \\ 
		22 & 0.495 & 0.249 & 0.495 & 0.746 & 0.495 & 0.830 & 0.495 & 0.125 & 8.746 \\ 
		23 & 0.495 & 0.746 & 0.495 & 0.249 & 0.495 & 0.830 & 0.495 & 0.125 & 8.746 \\ 
		24 & 0.495 & 0.249 & 0.495 & 0.249 & 0.495 & 0.166 & 0.495 & 0.873 & 8.753 \\ 
		25 & 0.495 & 0.746 & 0.495 & 0.746 & 0.495 & 0.498 & 0.495 & 0.374 & 9.075 \\ 
		26 & 0.495 & 0.249 & 0.495 & 0.746 & 0.495 & 0.498 & 0.495 & 0.623 & 9.080 \\ 
		27 & 0.495 & 0.746 & 0.495 & 0.249 & 0.495 & 0.498 & 0.495 & 0.623 & 9.080 \\ 
		28 & 0.495 & 0.746 & 0.495 & 0.746 & 0.495 & 0.166 & 0.495 & 0.623 & 9.741 \\ 
		29 & 0.495 & 0.249 & 0.495 & 0.746 & 0.495 & 0.830 & 0.495 & 0.374 & 9.743 \\ 
		30 & 0.495 & 0.746 & 0.495 & 0.249 & 0.495 & 0.830 & 0.495 & 0.374 & 9.743 \\ 
		31 & 0.495 & 0.249 & 0.495 & 0.249 & 0.495 & 0.830 & 0.495 & 0.623 & 9.748 \\ 
		32 & 0.495 & 0.249 & 0.495 & 0.249 & 0.495 & 0.498 & 0.495 & 0.873 & 10.082 \\ 
		33 & 0.495 & 0.746 & 0.495 & 0.746 & 0.495 & 0.830 & 0.495 & 0.125 & 10.735 \\ 
		34 & 0.495 & 0.249 & 0.495 & 0.746 & 0.495 & 0.166 & 0.495 & 0.873 & 10.743 \\ 
		35 & 0.495 & 0.746 & 0.495 & 0.249 & 0.495 & 0.166 & 0.495 & 0.873 & 10.743 \\ 
		36 & 0.495 & 0.746 & 0.495 & 0.746 & 0.495 & 0.498 & 0.495 & 0.623 & 11.070 \\ 
		37 & 0.495 & 0.746 & 0.495 & 0.746 & 0.495 & 0.830 & 0.495 & 0.374 & 11.733 \\ 
		38 & 0.495 & 0.249 & 0.495 & 0.746 & 0.495 & 0.830 & 0.495 & 0.623 & 11.738 \\ 
		39 & 0.495 & 0.746 & 0.495 & 0.249 & 0.495 & 0.830 & 0.495 & 0.623 & 11.738 \\ 
		40 & 0.495 & 0.249 & 0.495 & 0.746 & 0.495 & 0.498 & 0.495 & 0.873 & 12.072 \\ 
		41 & 0.495 & 0.746 & 0.495 & 0.249 & 0.495 & 0.498 & 0.495 & 0.873 & 12.072 \\ 
		42 & 0.495 & 0.746 & 0.495 & 0.746 & 0.495 & 0.166 & 0.495 & 0.873 & 12.733 \\ 
		43 & 0.495 & 0.249 & 0.495 & 0.249 & 0.495 & 0.830 & 0.495 & 0.873 & 12.740 \\ 
		44 & 0.495 & 0.746 & 0.495 & 0.746 & 0.495 & 0.830 & 0.495 & 0.623 & 13.728 \\ 
		45 & 0.495 & 0.746 & 0.495 & 0.746 & 0.495 & 0.498 & 0.495 & 0.873 & 14.062 \\ 
		46 & 0.495 & 0.249 & 0.495 & 0.746 & 0.495 & 0.830 & 0.495 & 0.873 & 14.730 \\ 
		47 & 0.495 & 0.746 & 0.495 & 0.249 & 0.495 & 0.830 & 0.495 & 0.873 & 14.730 \\ 
		48 & 0.495 & 0.746 & 0.495 & 0.746 & 0.495 & 0.830 & 0.495 & 0.873 & 16.720 \\ 
	\end{longtable}
\end{@empty}

\FloatBarrier
\clearpage
\subsection{\textsc{Shekel} Function}

\begin{table}[htbp]
	\centering
	\caption{Local Minima of \num{5}-Parametric \textsc{Shekel}}
	\label{tabAppMinimaShekel.5}
	\sisetup{round-mode = places, round-precision = 3, table-format = -2.3e0}
	\begin{tabular}{rSSSSS}
		\toprule
		\thead{\#} & \multicolumn{4}{c}{$\xvec$} & {\thead{$y$}}\\
		\midrule
		1 & 4.000 & 4.000 & 4.000 & 4.000 & -10.153 \\ 
		2 & 8.000 & 8.000 & 8.000 & 8.000 & -5.101 \\ 
		3 & 1.000 & 1.000 & 1.000 & 1.000 & -5.055 \\ 
		4 & 5.999 & 6.000 & 5.999 & 6.000 & -2.683 \\ 
		5 & 3.002 & 6.998 & 3.002 & 6.998 & -2.630 \\
		\bottomrule
	\end{tabular}
\end{table}

\begin{table}[htbp]
	\centering
	\caption{Local Minima of \num{7}-Parametric \textsc{Shekel}}
	\label{tabAppMinimaShekel.7}
	\sisetup{round-mode = places, round-precision = 3, table-format = -1.3e0}
	\begin{tabular}{rSSSSS[table-format = -2.3e0]}
		\toprule
		\thead{\#} & \multicolumn{4}{c}{$\xvec$} & {\thead{$y$}}\\
		\midrule
		1 & 4.000 & 4.000 & 4.000 & 4.000 & -10.403 \\ 
		2 & 8.000 & 8.000 & 8.000 & 8.000 & -5.129 \\ 
		3 & 1.000 & 1.000 & 1.000 & 1.000 & -5.088 \\ 
		4 & 4.995 & 3.006 & 4.995 & 3.006 & -3.703 \\ 
		5 & 5.998 & 5.999 & 5.998 & 5.999 & -2.752 \\ 
		6 & 3.001 & 7.001 & 3.001 & 7.001 & -2.750 \\ 
		7 & 2.005 & 8.992 & 2.005 & 8.992 & -1.833 \\	
		\bottomrule
	\end{tabular}
\end{table}

\begin{table}[htbp]
	\centering
	\caption{Local Minima of \num{10}-Parametric \textsc{Shekel}}
	\label{tabAppMinimaShekel.10}
	\sisetup{round-mode = places, round-precision = 3, table-format = -1.3e0}
	\begin{tabular}{rSSSSS[table-format = -2.3e0]}
		\toprule
		\thead{\#} & \multicolumn{4}{c}{$\xvec$} & {\thead{$y$}}\\
		\midrule
		1  & 4.000 & 4.000 & 4.000 & 4.000 & -10.536 \\ 
		2  & 8.000 & 8.000 & 8.000 & 8.000 & -5.176 \\ 
		3  & 1.000 & 1.000 & 1.000 & 1.000 & -5.128 \\ 
		4  & 5.002 & 3.002 & 5.002 & 3.002 & -4.070 \\ 
		5  & 5.999 & 5.997 & 5.999 & 5.997 & -2.871 \\ 
		6  & 3.001 & 7.000 & 3.001 & 7.000 & -2.790 \\ 
		7  & 5.992 & 2.022 & 5.992 & 2.022 & -2.608 \\ 
		8  & 6.986 & 3.593 & 6.986 & 3.593 & -2.495 \\ 
		9  & 2.005 & 8.991 & 2.005 & 8.991 & -1.854 \\ 
		10 & 7.985 & 1.013 & 7.985 & 1.013 & -1.696 \\ 
		\bottomrule
	\end{tabular}
\end{table}

\section{Contour Plots}
\label{secAppContourPlots}

Contour plots of the performance measurements peak ratio ($\PR$) and averaged \textsc{Hausdorff} distance ($\AHD$) over the number of initial and sequential design points for infill criteria $\GEILM$ are shown in this section.
Each contour line is smoothed by LOESS \parencite{cleveland_local_2017}.

\FloatBarrier
\subsection{Showcases}

\begin{figure}[htbp]
	\centering
	\includegraphics[page=2, trim=0mm 0mm 0mm 0mm, clip, width=\linewidth]{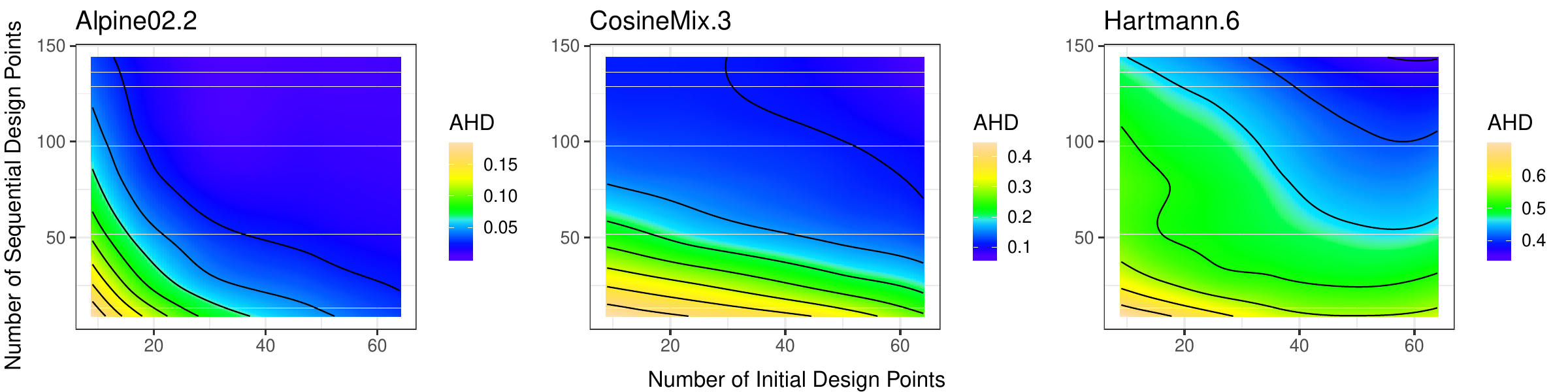}
	\caption{Contour Plot of $\AHD$}
	\label{figAppGEILMAHD}
\end{figure}

\FloatBarrier
\clearpage
\subsection{All Objective Functions}

\begin{figure}[htbp]
	\centering
	\includegraphics[trim=0mm 0mm 0mm 0mm, clip, width=\linewidth]{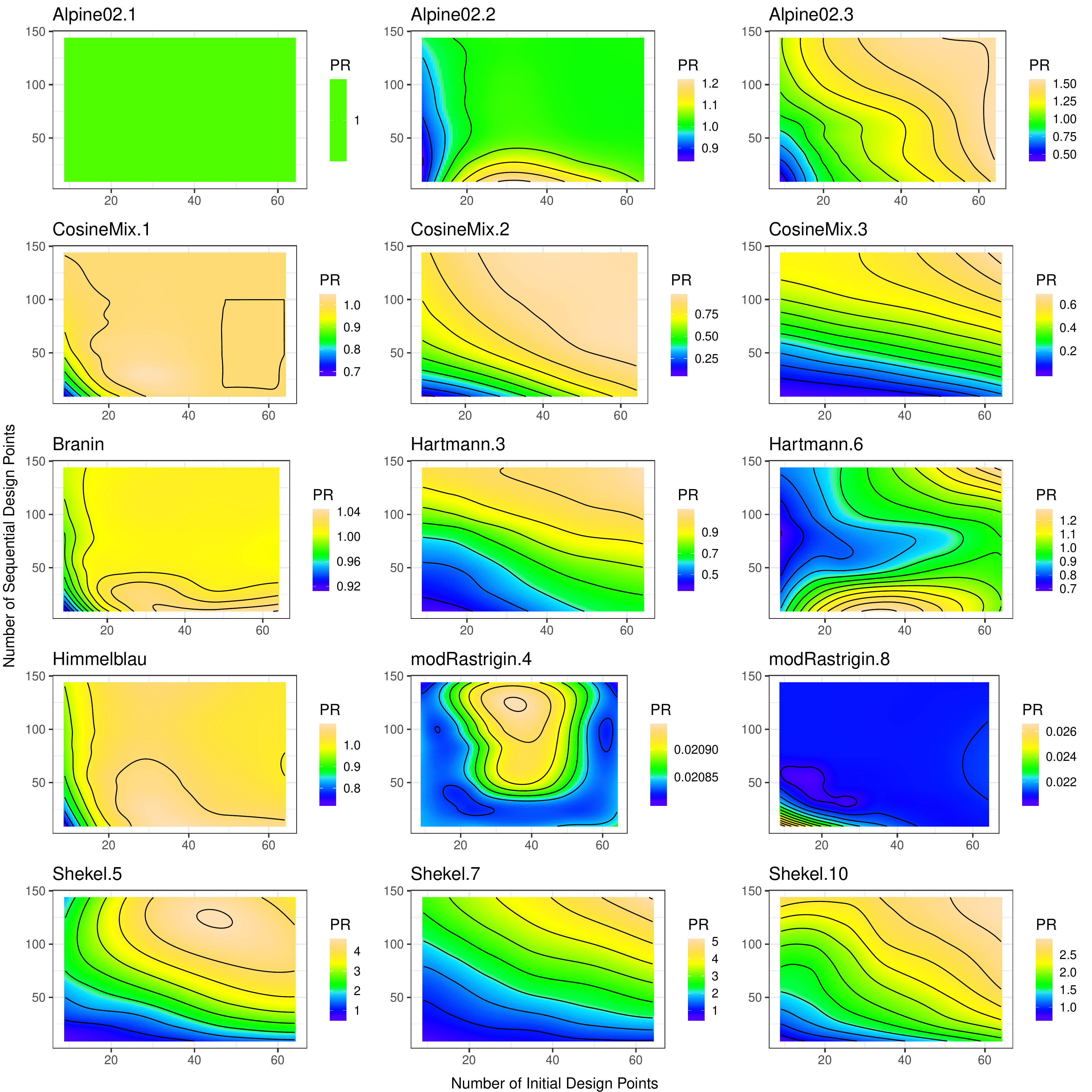}
	\caption{Contour Plot of $\PR$ for all Objective Functions}
	\label{figAppGEILMPRall}
\end{figure}

\begin{figure}[htbp]
	\centering
	\includegraphics[page=2, trim=0mm 0mm 0mm 0mm, clip, width=\linewidth]{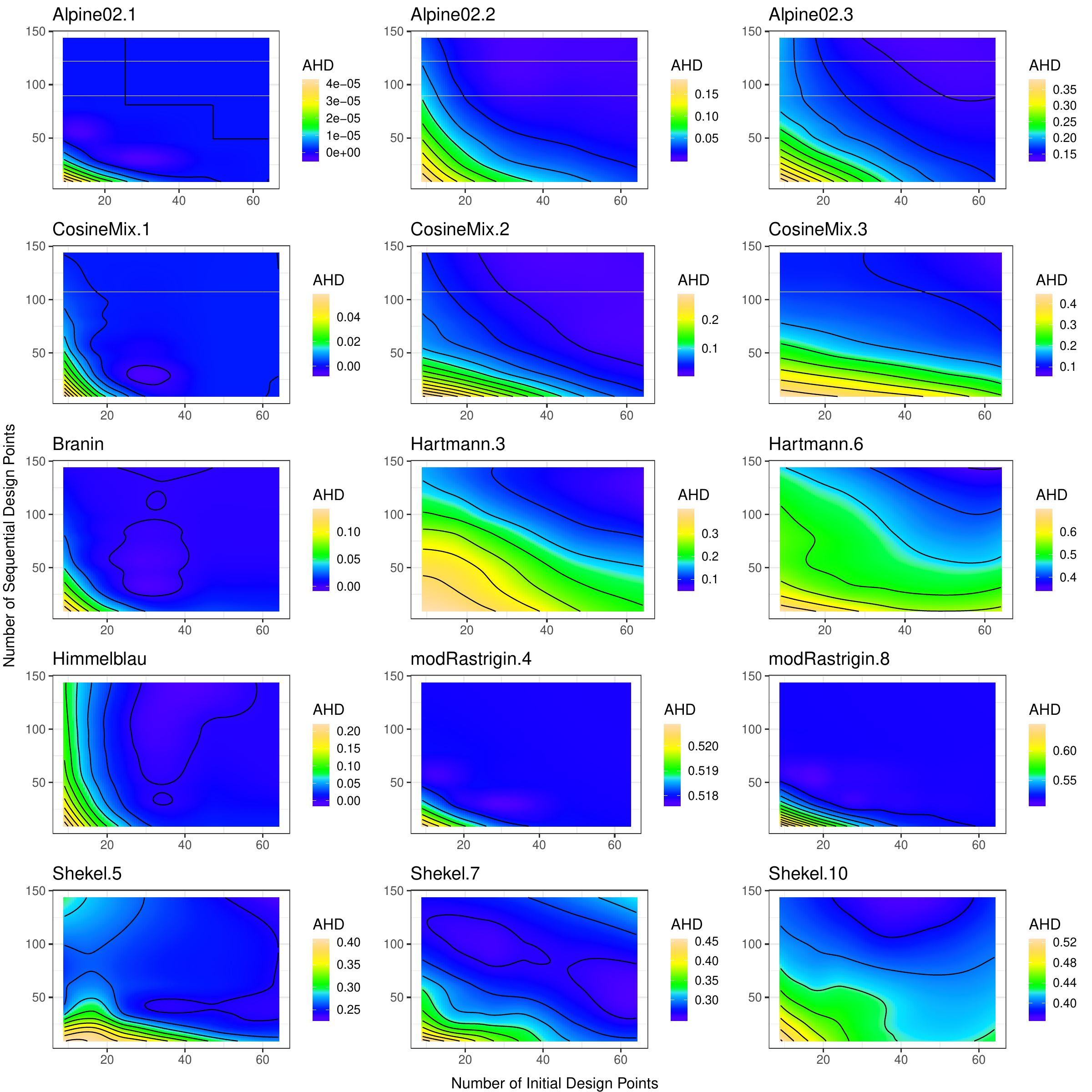}
	\caption{Contour Plot of $\AHD$ for all Objective Functions}
	\label{figAppGEILMAHDall}
\end{figure}

% References
\printbibliography

\end{document}